\documentclass[10pt,twocolumn,letterpaper]{article}

\usepackage[pagenumbers]{cvpr} 

\usepackage{booktabs}
\usepackage{multirow}
\usepackage{graphicx}
\usepackage{amsmath}
\usepackage{tabularx}
\usepackage{amssymb}
\usepackage{caption}
\usepackage{arydshln}
\usepackage[table]{xcolor}  
\usepackage{makecell}       

\newcommand{\ourrow}{\rowcolor{gray!7}}

\newcommand{\ci}[1]{\tiny{\textcolor{gray}{~($\pm #1$})}}

\definecolor{cvprblue}{rgb}{0.21,0.49,0.74}
\usepackage[pagebackref,breaklinks,colorlinks,allcolors=cvprblue]{hyperref}
\usepackage{enumitem}
\usepackage{multicol}
\usepackage{multirow}
\usepackage{graphicx}
\usepackage{xspace}
\usepackage{xcolor}
\usepackage{caption}
\usepackage{bbm}
\usepackage{dsfont}
\usepackage{mmstyle}
\usepackage{wrapfig}
\usepackage{bbding}  
\usepackage{pifont}  
\usepackage[capitalize]{cleveref}
\crefname{section}{Sec.}{Secs.}
\Crefname{section}{Section}{Sections}
\Crefname{table}{Table}{Tables}
\crefname{table}{Tab.}{Tabs.}
\crefname{appendix}{App.}{Apps.}
\Crefname{appendix}{Appendix}{Appendices}

\usepackage{booktabs}
\usepackage{float}
\usepackage{amsmath}  
\usepackage{amssymb}  
\usepackage{bm}
\usepackage{dsfont}
\setlist[enumerate]{label=\arabic*., leftmargin=*, itemsep=0.25em}
\newcommand{\ourshort}{\textbf{Gallant}\xspace}

\def\paperID{4020} 
\def\confName{CVPR}
\def\confYear{2026}

\title{Gallant: Voxel Grid-based Humanoid Locomotion and \\Local-navigation across 3D Constrained Terrains}

\author{
Qingwei Ben$^{*,2,1}$, Botian Xu$^{*,2}$, Kailin Li$^{*,1}$, Feiyu Jia$^{1,3}$,\\
Wentao Zhang$^{4,1}$, Jingping Wang$^{1,5}$, Jingbo Wang$^{1}$, Dahua Lin$^{2,1}$, Jiangmiao Pang$^{\oplus,1}$\\
{\small $^{1}$Shanghai Artificial Intelligence Laboratory, $^{2}$The Chinese University of Hong Kong, }\\
{\small $^{3}$University of Science and Technology of China
, $^{4}$University of Tokyo, $^{5}$Shanghai Jiaotong University}\\
{\small $^{*}$Equal Contribution, $^{\oplus}$Corresponding Author}
}
\definecolor{mycolor}{rgb}{0.07,0.419,0.682}
\hypersetup{
    colorlinks=true,
    linkcolor=mycolor, 
    urlcolor=mycolor,  
    citecolor=mycolor, 
}
           
\newcommand{\insertteaser}{%
  \begin{flushleft}
    \centering
    \vspace{-30pt}
    \includegraphics[clip,trim=0cm 0cm 0cm 0cm,width=0.98\textwidth]{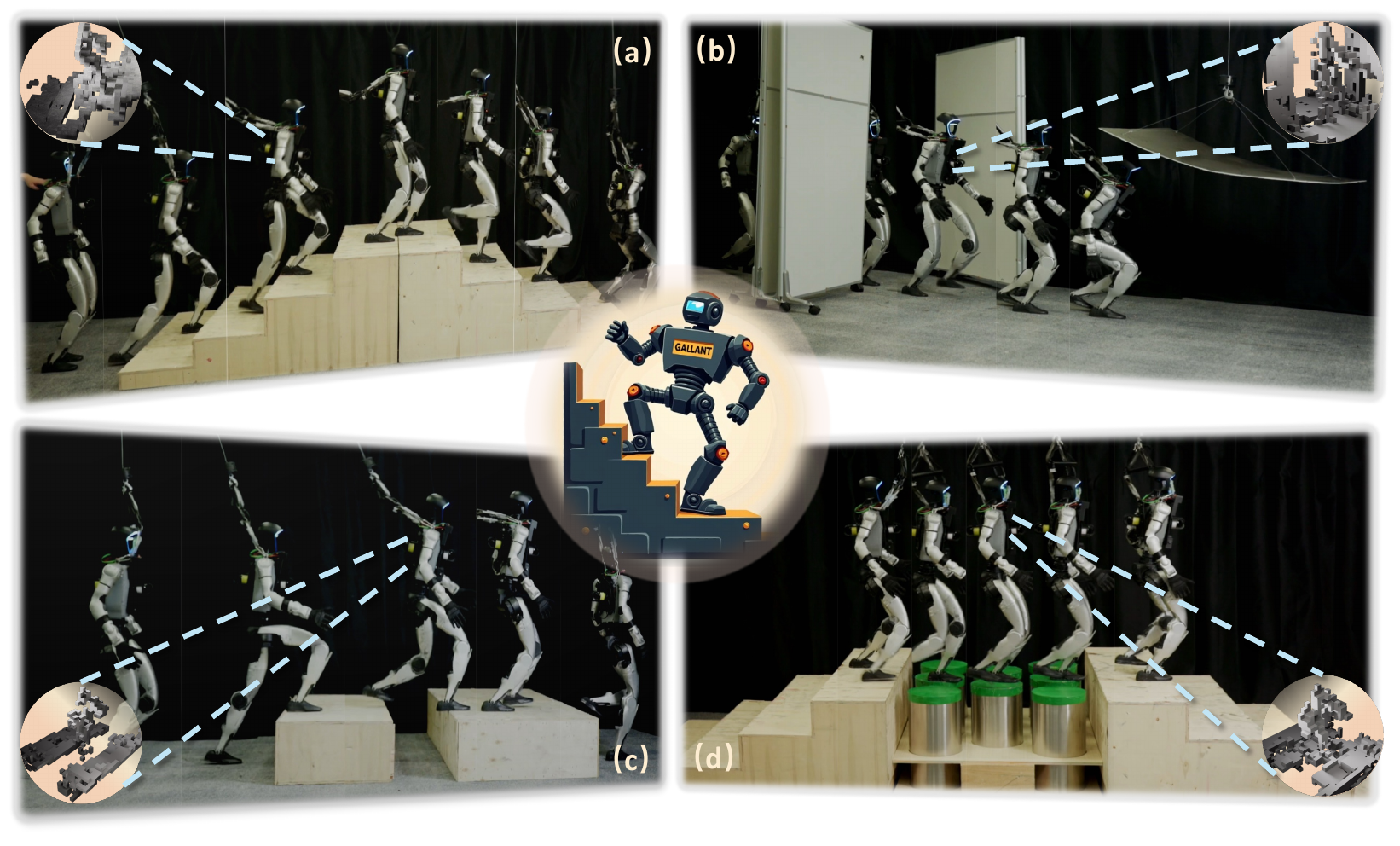}%
    \captionsetup{justification=raggedright,singlelinecheck=false}
    \vspace{-11pt}
    \captionof{figure}{
      \textbf{Overview.} Gallant enables a single policy with voxel grids to traverse diverse 3D constrained terrains: (a) ascend and descend stairs, (b) pass doors and duck under ceilings, (c) step onto platforms and over gaps, and (d) cross stepping-stone pillars.
    }
    \label{fig:teaser}
  \end{flushleft}
}
\makeatletter
\newif\ifteaser@used
\teaser@usedfalse

\let\old@maketitle\@maketitle
\renewcommand{\@maketitle}{%
  \old@maketitle
  \ifteaser@used\relax\else
    \teaser@usedtrue
    \insertteaser
  \fi
}
\makeatother
\begin{document}
\maketitle
\begin{abstract}
Robust humanoid locomotion requires accurate and globally consistent perception of the surrounding 3D environment. However, existing perception modules, mainly based on depth images or elevation maps, offer only partial and locally flattened views of the environment, failing to capture the full 3D structure.
This paper presents \textbf{Gallant}, a voxel-grid–based framework for humanoid locomotion and local navigation in 3D constrained terrains.
It leverages voxelized LiDAR data as a lightweight and structured perceptual representation, and employs a z-grouped 2D CNN to map this representation to the control policy, enabling fully end-to-end optimization. A high-fidelity LiDAR simulation that dynamically generates realistic observations is developed to support scalable, LiDAR-based training and ensure sim-to-real consistency.
Experimental results show that Gallant’s broader perceptual coverage facilitates the use of a single policy that goes beyond the limitations of previous methods confined to ground-level obstacles, extending to lateral clutter, overhead constraints, multi-level structures, and narrow passages. Gallant also firstly achieves near-100\% success rates in challenging scenarios such as stair climbing and stepping onto elevated platforms through improved end-to-end optimization. Website: \href{https://gallantloco.github.io/}{Gallant}.
\end{abstract}

\newpage
\section{Introduction}
\label{sec:intro}
Robust humanoid locomotion in unstructured 3D environments demands accurate and globally consistent perception of surrounding geometry. While recent systems have progressed from lab prototypes to real-world deployment~\citep{intelligence2504pi0,generalist2025gen0,luo2025sonic}, ensuring operational safety remains a key challenge. Robots must not only traverse level surfaces, but also navigate terrain irregularities, ground-level obstacles, lateral clutter, and overhead constraints. This requires a perception architecture that enables anticipatory collision checking, clearance-aware motion generation, and planning of contact-rich maneuvers.

Existing perception modules, such as those based on depth images or elevation maps, provide only partial and locally flattened views of the environment, limiting the robot's understanding of complex 3D structures. Depth cameras\citep{zhuang2024humanoid,sun2025dpl} offer lower‑latency perception; however, their narrow field of view (FoV) and limited range impede reasoning about complex, spatially extended environments. In contrast, 3D LiDAR provides detailed scene geometry with a wide FoV, but its raw point clouds are sparse and noisy, which bottlenecks sample‑efficient policy learning and real‑time inference. Elevation‑mapping approaches compress full 3D LiDAR point clouds into 2.5D height fields\citep{long2025learning,wang2025beamdojo,ren2025vb,he2025attention,videomimic}, yielding a bird’s‑eye height estimate for each ground‑plane cell~\citep{Fankhauser2018ProbabilisticTerrainMapping,Fankhauser2014RobotCentricElevationMapping}. This projection discards vertical and multilayer structure (e.g., overhangs, low ceilings, mezzanines, stair undersides), and the reconstruction stage can introduce algorithm‑specific distortions and latency, further decoupling perception from control.

To address these limitations, we introduce \textbf{Gallant}, a voxel-\textbf{g}rid–based perception-learning framework for hum\textbf{a}noid \textbf{l}ocomotion and \textbf{l}oco-navigation \textbf{a}cross 3D co\textbf{n}strained \textbf{t}errains. Gallant uses a robot-centric voxel grid derived from LiDAR point clouds as its perception representation, preserving multi-layer scene structure over a large FoV while aggregating raw points into voxels to reduce dimensionality and smooth noise, yielding a lightweight tensor amenable to efficient learning. A \emph{z-grouped} 2D convolutional neural network (CNN) treats height slices as channels, exploiting sparsity to produce compact features with a favorable accuracy–compute trade-off compared to heavier 3D CNNs~\citep{maturana2015voxnet,tran2015learning,frey2022locomotion}. These features are fused with proprioceptive signals and passed to a multi-layer perceptron (MLP)-based actor for whole-body control, directly conditioning actions on 3D structural cues. To scale training and narrow the simulation-to-reality (sim-to-real) gap, we develop a LiDAR simulation pipeline that models sensor noise and latency and enables realistic scanning of dynamic objects, including the robot’s own moving links, thereby aligning synthetic data with deployment conditions. We further construct eight representative terrain families spanning ground-level obstacles, lateral clutter, and overhead constraints, encouraging the policy to internalize structural regularities critical for generalization.

Experimental results show that Gallant enables a single end-to-end policy that generalizes from simulation to diverse real-world environments, handling not only ground-level obstacles but also lateral clutter and overhead constraints—capabilities that were previously beyond the reach of existing methods. Gallant achieves near-100\% success in challenging tasks such as stair climbing and platform stepping, while significantly improving robustness over elevation-based baselines. Experiments also highlight the importance of our high-fidelity LiDAR simulation, which dynamically generates realistic observations essential for scalable, LiDAR-based training. Ablation results further demonstrate the efficiency of the z-grouped 2D CNN, which attains superior performance and lower inference latency compared to 3D CNNs, making it well-suited for real-time humanoid deployment. These results establish Gallant as a practical, full-stack solution—from realistic LiDAR simulation to robust control—for full-space perceptive locomotion and local-navigation in 3D constrained environments. Our contribution lies in the following aspects:

\begin{enumerate}
\item We propose voxel grid as a lightweight yet geometry-preserving representation for humanoid locomotion and loco-navigation~\citep{rudin2022advanced} in 3D-constrained environments.
\item We verify that $z$-grouped 2D CNN effectively processes voxel grids, offering a favorable trade-off between representation capacity and computational efficiency.
\item We develop a full-stack pipeline from sensor simulation to policy training, achieving a single policy that generalizes across diverse 3D-constrained terrains in real.
\end{enumerate}

\section{Related Work}
\label{sec:related_work}

\begin{table}[!ht]
    \centering
    \vspace{-5pt}
    \caption{Comparison between gallant and previous methods. FoV in \textit{Solid Angles} are computed by parameter of the used sensors.}
    \resizebox{1\linewidth}{!}{
    \begin{tabular}{lccccc}
    \toprule 
    \multirow{2}{*}{Method} & \multirow{2}{*}{\shortstack{Perceptual\\Representation}} & \multirow{2}{*}{\shortstack{Fov}} &\multirow{2}{*}{\shortstack{Ground}} & \multirow{2}{*}{\shortstack{Lateral}} & \multirow{2}{*}{\shortstack{Overheading}} \\ \\
    \midrule
    \citet{long2025learning} & Elevation Map & $\sim 1.97\pi$ & \Checkmark & \XSolidBrush & \XSolidBrush  \\
    \citet{wang2025beamdojo} & Elevation Map &  $\sim 1.97\pi$ &\Checkmark & \XSolidBrush &  \XSolidBrush  \\
    \citet{ren2025vb} & Elevation Map & $\sim 1.97\pi$ &\Checkmark & \Checkmark & \XSolidBrush  \\ 
    \citet{zhuang2024humanoid} & Depth Image & $\sim 0.43\pi$ &\Checkmark & \XSolidBrush & \XSolidBrush  \\ 
    \citet{wang2025omni} & Point Cloud & $\sim 1.97\pi$ &\XSolidBrush & \Checkmark & \Checkmark \\ 
    \cmidrule(r){1-6}
    \ourshort (ours) & Voxel Grid & $\sim 4.00\pi$ &\Checkmark & \Checkmark & \Checkmark\\

    \bottomrule
    \end{tabular}}
    \label{table:comparision_method}
\end{table}
\textbf{Humanoid Perceptive Locomotion.} Humanoid perceptive locomotion uses onboard sensing to traverse constrained terrains. Prior work mainly relies on elevation maps~\citep{long2025learning,ren2025vb,wang2025beamdojo,sun2025learning,miki2022elevation}, trained with ground-truth height fields and deployed via LiDAR reconstruction~\cite{Fankhauser2018ProbabilisticTerrainMapping,Fankhauser2014RobotCentricElevationMapping}. While effective for ground-level reasoning, elevation maps flatten the scene and ignore lateral or overhead structures, and introduce reconstruction latency. Alternatively, depth cameras offer higher update rates and are proved to be effective on quadruped robots ~\cite{zhuang2024humanoid,sun2025dpl,cheng2024extreme,li2025move,loquercio2023learning,agarwal2023legged,zhuang2023robot}, but their narrow field of view and limited spatial continuity similarly restrict 3D understanding, hindering policy generalization in diverse environments. Recent LiDAR simulation advances enable realistic sensing during training. While point-cloud–based inputs~\cite{wang2025omni,hoeller2024anymal} address prior limitations, their high processing cost makes real-time onboard use infeasible. Voxel grids offer a structured, efficient alternative~\cite{frey2022locomotion,niijima2025real}. They have been explored for cross-modal perception for scene-understanding on legged robots~\cite{shi2025oneocc,zhang2025humanoidpano,cui2025humanoid}, but remain unused in humanoid locomotion as a direct way of perception. To this end, Gallant introduces a LiDAR perception framework tailored for scalable simulation and real-time deployment, using voxel grids to support a single policy capable of zero-shot sim-to-real transfer and full 3D obstacle handling. A comparison of Gallant with prior methods in perceptual representation, FoV, and supported obstacle types is listed in \cref{table:comparision_method}.

\textbf{Local Navigation.}
Local navigation enables legged robots to reach targets in cluttered, constrained environments while minimizing incidental contact. Most systems adopt a hierarchical design: a high-level planner outputs velocity commands, and a low-level policy tracks them~\cite{frey2022locomotion,yokoyama2023asc,zhang2024resilient,cheng2024navila,wei2025streamvln,cai2025navdp,liu2024visual,qiu2025wildlma,hoeller2024anymal}. This decoupling limits the policy’s ability to exploit terrain, and tracking errors—combined with slow high-level updates, further degrading performance. Recent work explores end-to-end training by adding obstacle-avoidance rewards to velocity tracking~\cite{ren2025vb}, but this creates conflicting objectives. Using target positions instead allows the policy to reason over terrain and choose appropriate actions~\cite{rudin2022advanced,he2024agile,yang2021learning}, though this remains untested on humanoids. Gallant adopts this position-based formulation to fuse local navigation and locomotion into one single policy.

\section{Method}
\label{sec:method}
\begin{figure*}[!ht]
  \centering
  \includegraphics[width=1.00\textwidth]{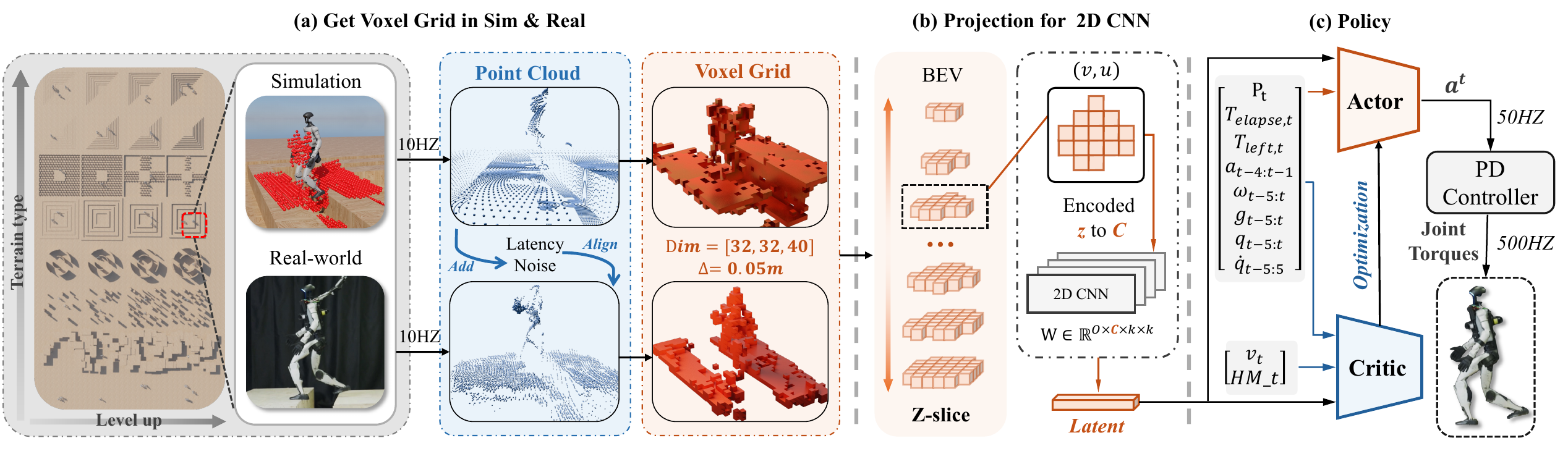}
  \caption{\textbf{Method Overview.} (a) Curriculum-based training over 8 representative terrains enhances generalization. (b) Realistic voxel path alignment achieved via efficient LiDAR simulation with domain-randomized latency and noise. (c) A 2D CNN-based perceptual module processes voxel grid using the z-dimension as input channels, balancing efficiency and representation capability. (d) A latent-aware PPO policy enables zero-shot sim-to-real transfer across diverse obstacles, including ground, lateral, and overhead challenges.}
  \label{fig:overview}
  \vspace{-10pt}
\end{figure*}
We introduce \ourshort, a voxel-grid–based perceptive learning framework for humanoid locomotion and local navigation~\citep{rudin2022advanced} in 3D constrained environments. As shown in \cref{fig:overview}, the system comprises: (i) a parallelized LiDAR simulation pipeline (\cref{sec:method_lidar}), (ii) a lightweight 2D CNN perception module tailored to sparse voxel grids (\cref{sec:method_cnn}), and (iii) a set of representative terrain families for curriculum training (\cref{sec:method_terrain}). Together, these components form a full-stack pipeline—from data generation to perception to control—that trains a single policy to robustly traverse all-space obstacles and deploy zero-shot on real hardware.

\subsection{Problem Formulation}
We formulate humanoid perceptive locomotion as a partially observable Markov decision process (POMDP) $\mathcal{M} = (\mathcal{S}, \mathcal{A}, \mathcal{O}, P, \mathcal{R}, \Omega, \gamma)$ and train an actor–critic policy using Proximal Policy Optimization (PPO)~\cite{schulman2017proximal}. The training environment is divided into \(8\,\mathrm{m} \times 8\,\mathrm{m}\) blocks. At each episode, the humanoid starts at the block center, and a goal $\mathbf{G}$ is sampled along the perimeter, with a fixed horizon of 10 seconds for robots to reach. The observation at time $t$ is defined as:
\vspace{-3pt}
\begin{align*}
o_t = (&\,\underbrace{\mathbf{P}_t,\, \mathbf{T}_{\text{elapse},t},\, \mathbf{T}_{\text{left},t}}_{\text{Command}},\;
\underbrace{a_{t-4:t-1}}_{\text{Action history}},\\
&\,\underbrace{\omega_{t-5:t},\, g_{t-5:t},\, q_{t-5:t},\, \dot{q}_{t-5:t}}_{\text{Proprioception}},\;\\&
\underbrace{\texttt{Voxel\_Grid}_t}_{\text{Perception}},\;
\underbrace{v_t,\, \texttt{Height\_Map}_t}_{\text{Privileged}}),
\end{align*}
where: $\mathbf{P}_t$ is the goal position relative to the robot base, $\mathbf{T}_{\text{elapse},t}$ is the elapsed time in the episode, $\mathbf{T}_{\text{left},t} = T - \mathbf{T}_{\text{pass},t}$ is the remaining time until timeout ($T=10$s), $a_{t}$ denotes actions output by policy, $\omega_t$ and $v_t$ are the root angular and linear velocity of the robot, $g_t$ is the vector $[0, 0, -1]$ projected into the robot base frame, $q_t$ and $\dot{q}_t$ are joint positions and velocities, respectively, $\texttt{Voxel\_Grid}_t$ is the voxelized perception input, $\texttt{Height\_Map}_t$ is relative heights of the scanned dots to the robot. Here, the subscript range $t-a:t-b$ denotes inclusion of temporal history from time step $t-a$ to $t-b$. Actor and critic share all features except privileged inputs, which are critic-only. The reward follows~\citet{ben2025homie} with velocity tracking rewards replaced by goal-reaching reward~\cite{rudin2022advanced}:
\[
r_{\text{reach}} = \frac{1}{1 + \|\mathbf{P}_t\|^2} \cdot \frac{\mathds{1}(t > T - T_r)}{T_r} \,\,(T_r=2s),
\]
allowing time for trajectory exploration. The objective is to maximize expected return $J(\pi) = \mathbb{E}[\sum_{t=0}^{H-1} \gamma^t r_t]$. Episodes end on fall, harsh collision, or timeout. Network and reward details are listed in Appendix.

\subsection{Efficient LiDAR Simulation}
\label{sec:method_lidar}
Most GPU-based simulators, such as IsaacGym and IsaacSim, either lack native support for efficient LiDAR simulation or are limited to scanning a single static mesh. However, realistic simulation in dynamic environments requires accounting for all relevant geometry, including both static and dynamic meshes—especially bodies of robots. To address this, we implement a lightweight, efficient raycast-voxelization pipeline using NVIDIA Warp~\cite{warp2022}. Traditional raycasting builds a Bounding Volume Hierarchy (BVH) over scene geometry, which becomes costly if updated at every simulation step due to dynamics. To mitigate this, we precompute a BVH for each mesh in its local (body) frame. During simulation, the ray origin $\mathbf{p}$ is transformed, and only the rotation component is applied to the direction $\mathbf{d}$, rotating the ray into the mesh’s local frame. To be specific, the raycasting is performed as:
\begin{equation*}
\text{raycast}(TM, \mathbf{p}, \mathbf{d}) = T^{-1} \text{raycast}(M, T^{-1}\mathbf{p}, R^{-1}\mathbf{d}),
\end{equation*}
Here, $T$ denotes the full transformation matrix, and $R$ its rotational component. The function returns the ray–mesh intersection point. At each simulation step, ray–mesh intersections are computed for every mesh $M$ using its transform $T_t$, parallelized via a Warp kernel of shape $(N_{\text{envs}}, N_{\text{meshes}}, N_{\text{rays}})$. Rays are emitted from the LiDAR origin $P_{\text{LiDAR}}$ in directions defined as $O_{\text{ray}_i} = O_{\text{LiDAR}} + O_{\text{ray}_i,\text{offset}}$, where $O_{\text{ray}_i,\text{offset}}$ is the $i$-th ray’s direction offset from the LiDAR orientation. Let $P_i$ be the hit position of the $i$-th ray; the resulting point cloud is: $\mathcal{P}_t = \bigcup_{i=1}^{N_{\text{rays}}} \left\{ P_i \right\}$,
which is subsequently used to construct the voxel grid.
To align simulation with real-world sensing, we apply domain randomization:
\textit{(a) LiDAR Pose:} Perturbed at episode start by $P_{\text{LiDAR}}^{\text{rand}} = P_{\text{LiDAR}} + \mathcal{N}(0,1)$ (cm) and $O_{\text{ray}_i}^{\text{rand}} = O_{\text{LiDAR}} + \mathcal{N}(0,(\frac{\pi}{180})^2) + O_{\text{ray}_i,\text{offset}}$ (rad); \textit{(b) Hit Position:} $P_i^{\text{rand}} = P_i + \mathcal{N}(0,1)$ (cm);  
\textit{(c) Latency:} Simulated at 10 Hz with 100–200 ms delay;  
\textit{(d) Missing Grid:} Randomly mask 2\% of voxels to model real-world dropout. These augmentations reduce the sim-to-real gap and improve policy transferability.

\subsection{Voxel Representation and 2D CNN Perception}
\label{sec:method_cnn}
We convert LiDAR point clouds into a fixed-size, robot-centric voxel grid. At each timestep, returns from two torso-mounted LiDARs are transformed into a unified torso frame. The perception volume is defined as a cuboid $\Omega = [-0.8, 0.8]\,\mathrm{m} \times [-0.8, 0.8]\,\mathrm{m} \times [-1.0, 1.0]\,\mathrm{m}$, discretized at resolution $\Delta = 0.05\,\mathrm{m}$, yielding a $32 \times 32 \times 40$ grid along the $x$, $y$, and $z$ axes respectively. Each voxel is set to 1 if at least one LiDAR point lies inside its volume, and 0 otherwise, producing a binary occupancy tensor $X \in \{0,1\}^{C \times H \times W}$, where $C = 40$ (height slices), $H = W = 32$ (spatial resolution).

Due to the line-of-sight nature of LiDAR and the structured nature of typical terrains, the voxel grid is highly sparse and locally concentrated: most $(x,y)$ columns contain only one or two occupied $z$-slices, and large contiguous spatial regions may remain completely empty. Rather than applying computationally expensive 3D convolutions over the full volume, we treat the $z$-axis as the channel dimension and apply 2D convolutions over the $x-y$ plane. This leverages spatial context while using channel mixing to capture vertical structure, making effective use of the sparse, localized occupancy pattern. Formally, let $X \in \mathbb{R}^{C \times H \times W}$ be the voxel input and $\mathbf{W} \in \mathbb{R}^{O \times C \times k \times k}$ the weights of a 2D convolution. The output $Y \in \mathbb{R}^{O \times H \times W}$ is computed:
\begin{equation*}
Y_{o,v,u} = \sigma \left(
\sum_{c=0}^{C-1} \sum_{\Delta v, \Delta u}
\mathbf{W}_{o,c,\Delta v,\Delta u} \cdot X_{c, v+\Delta v, u+\Delta u} + b_o
\right),
\end{equation*}
where $\sigma$ is a nonlinearity and $b_o$ is a bias term. Compared to a 3D kernel of size $k^3$, this design reduces compute and memory cost by roughly a factor of $k$, while still capturing the vertical patterns critical for locomotion. Moreover, the 2D structure enables efficient parallel training and supports real-time inference on onboard compute.

\subsection{Terrain Design}

We design 8 representative terrain types to train robots in simulation: 
\textbf{Plane} represents the easiest terrain and helps robots learn to walk in the early stage;  
\textbf{Ceiling} with randomized height and density requires reasoning about overhead constraints and crouching;  
\textbf{Forest}, composed of randomly spaced cylindrical pillars, represents sparse lateral clutter requiring weaving behavior;  
\textbf{Door} presents narrow gaps demanding precise lateral clearance;  
\textbf{Platform} consists of high, ring-shaped structures with variable spacing and height, requiring recognition of stepable surfaces and inter-platform traversal;  
\textbf{Pile} introduces fine-grained support reasoning for safe foot placement;  
\textbf{Upstair} and \textbf{Downstair} require continuous adaptation to vertical elevation.
\begin{table}[t]
\centering
\caption{Parameters for generating curriculum training terrains.}
\resizebox{0.98\linewidth}{!}{%
\begin{tabular}{llll}
\toprule Terrain Type $\tau$ & Term & $\mathbf{p}_\tau^{\min}$ & $\mathbf{p}_\tau^{\max}$ \\ 
\midrule 
Ceiling & Ceiling height ($m$) $\downarrow$ & 1.30 & 1.00
\\ & Number of Ceiling (-) $\uparrow$ & 10 & 40 \\ 
\cdashline{1-4} \noalign{\vskip 0.6mm}
Forest & Minimum distance between trees ($m$) $\downarrow$ & 2.0 & 1.0 \\
& Number of trees (-) $\uparrow$ & 3 & 32\\
\cdashline{1-4} \noalign{\vskip 0.6mm}
Door & Distance between two walls ($m$) $\downarrow$ & 2.00 & 1.00\\
& Width of the doors ($m$) $\downarrow$ & 1.60 & 0.80 \\
\cdashline{1-4} \noalign{\vskip 0.6mm}
Platform & Height of the platforms ($m$) $\uparrow$ & 0.05 & 0.35\\
& Gap width between two platforms ($m$) $\uparrow$ & 0.20 & 0.50 \\
\cdashline{1-4} \noalign{\vskip 0.6mm}
Pile & Distance between two cylinders ($m$) $\uparrow$ & 0.35 & 0.45 \\
\cdashline{1-4} \noalign{\vskip 0.6mm}
Upstair & Height of each step ($m$) $\uparrow$ & 0.00 & 0.20\\
& Width of each step ($m$) $\downarrow$ & 0.50 & 0.30 \\
\cdashline{1-4} \noalign{\vskip 0.6mm}
Downstair & Height of each step ($m$) $\uparrow$ & 0.00 & 0.20\\
& Width of each step ($m$) $\downarrow$ & 0.50 & 0.30 \\
\bottomrule 
\end{tabular}}
\label{table:para_terrain}
\end{table}
\begin{figure}[!ht]
    \centering
    \includegraphics[width=0.9\linewidth]{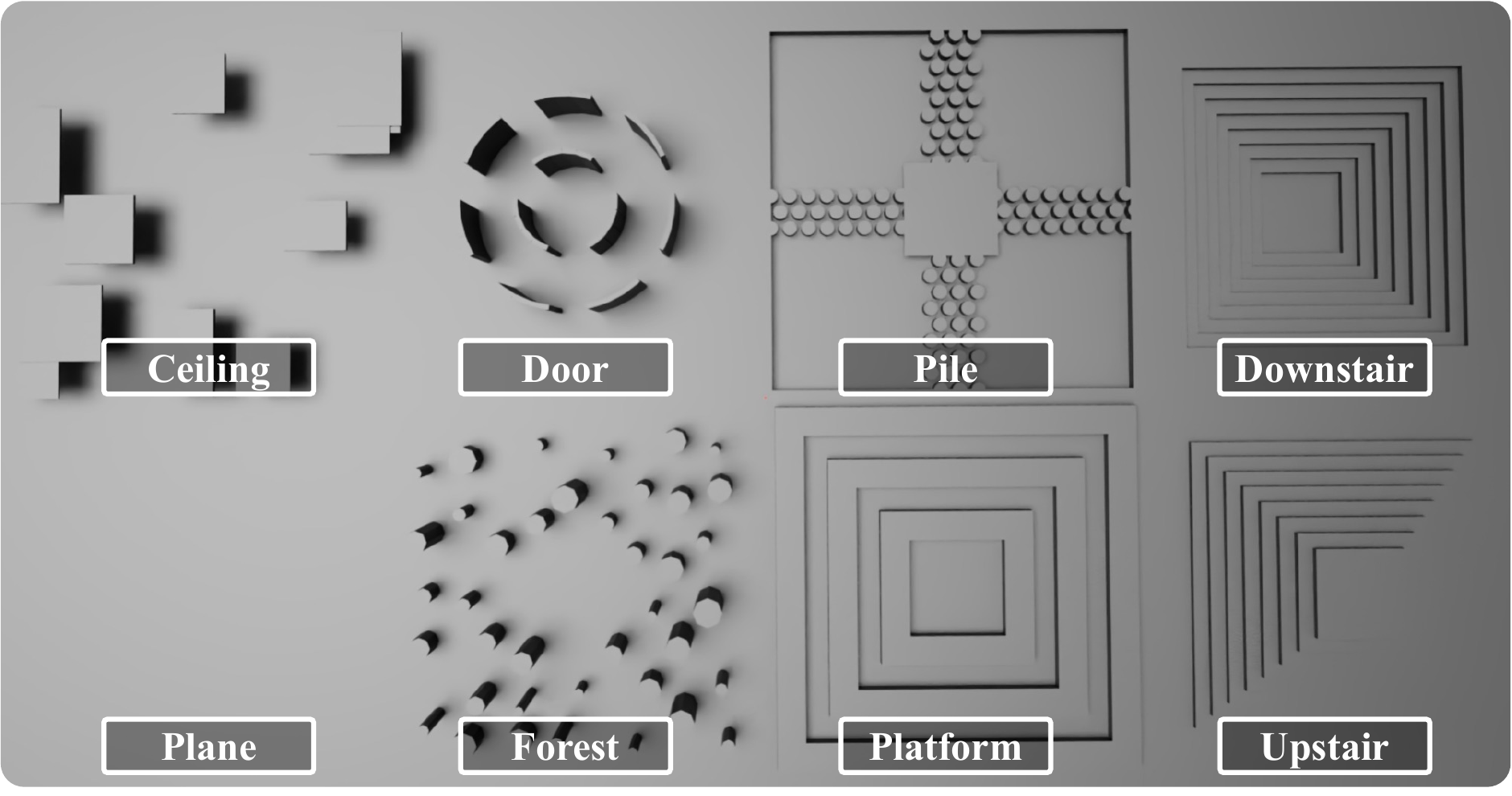}
    \caption{Terrain types used to train robots in simulation($\mathbf{p}_\tau^{\max}$)}
    \label{fig:terrain}
\end{figure}
We adopt a curriculum-based training strategy where terrain difficulty increases progressively. Each terrain type $\tau$ is parameterized by a scalar difficulty $s \in [0, 1]$. Terrain generation parameters are interpolated as:  
\[
\mathbf{p}_\tau(s) = (1-s)\,\mathbf{p}_\tau^{\min} + s\,\mathbf{p}_\tau^{\max},
\]
where $\mathbf{p}_\tau^{\min}$ and $\mathbf{p}_\tau^{\max}$ denote the easiest and hardest settings (see \cref{table:para_terrain}). In each episode, a 10s goal-reaching task is assigned, and success results in promotion to harder settings; failure leads to demotion. To support learning on \textbf{Pile}, we overlay a flat surface during early training (low $s$), following \cite{wang2025beamdojo}, allowing the robot to first learn basic foothold placement. For high $s$, the plane is removed, and training continues on fully gapped terrain for true crossing behavior.

\section{Experiments}
\label{sec:experiment}

\begingroup
\setlength{\tabcolsep}{3.5pt}
\begin{table*}[t]
    \centering
    \caption{\textbf{Simulation ablation results.} We present a success rate comparison between \ourshort and baselines on the eight representative terrains. The means and standard variation are reported across 5 evaluations, each with 1,000 testing episodes. Success rate is reported as a percentage (e.g., 90 means 90\%). For each ablation setting, the best-performing value per metric on each terrain is highlighted in bold.}
    
    \resizebox{1\linewidth}{!}{%
 \begin{tabular}{lc c cc c cc c cc c cc c cc c cc c cc c cc} 
 \toprule
  \multirow{2}{*}{\textbf{Method}} & & \multicolumn{2}{c}{Plane} & & \multicolumn{2}{c}{Ceiling} & & \multicolumn{2}{c}{Forest} & & \multicolumn{2}{c}{Door} & & \multicolumn{2}{c}{Platform} & & \multicolumn{2}{c}{Pile} & & \multicolumn{2}{c}{Upstair} & & \multicolumn{2}{c}{Downstair} \\ 
  \cmidrule{3-4}\cmidrule{6-7}\cmidrule{9-10} \cmidrule{12-13} \cmidrule{15-16}\cmidrule{18-19}\cmidrule{21-22} \cmidrule{24-25} 
    & 
    & $E_{\mathrm{succ}}\uparrow$  & $E_{\mathrm{collision}}\downarrow$ & &$E_{\mathrm{succ}}\uparrow$  & $E_{\mathrm{collision}}\downarrow$ 
    & 
    
    & $E_{\mathrm{succ}}\uparrow$  & $E_{\mathrm{collision}}\downarrow$  & & $E_{\mathrm{succ}}\uparrow$  & $E_{\mathrm{collision}}\downarrow$ 
    &
    
    & $E_{\mathrm{succ}}\uparrow$  & $E_{\mathrm{collision}}\downarrow$  & &$E_{\mathrm{succ}}\uparrow$  & $E_{\mathrm{collision}}\downarrow$ 
    &
    
    & $E_{\mathrm{succ}}\uparrow$  & $E_{\mathrm{collision}}\downarrow$  & &$E_{\mathrm{succ}}\uparrow$  & $E_{\mathrm{collision}}\downarrow$ 
    \\ 
 \midrule 
  \ourrow \multicolumn{21}{l}{\textbf{(a) Ablation on Self-scan}} & \\
  \cdashline{1-26}\noalign{\vskip 0.6mm}
w/o-Self-Scan & 
  & 99.7\ci{0.1} & 1.6\ci{3.2}& & 28.4\ci{2.4} & 442.7\ci{22.1} & 
  & 78.1\ci{1.4} & 420.5\ci{12.1}& & 98.3\ci{0.7} & 152.7\ci{20.0} & 
  & 22.16\ci{1.2} & 637.6\ci{31.3}& & 27.2\ci{1.0} & 579.0\ci{55.1} &
  & 33.0\ci{0.9} & 305.5\ci{16.6}& & 96.6\ci{0.4} & 15.15\ci{6.1} \\
\ourshort & 
 & \textbf{100.0\ci{0.0}} & \textbf{0.0\ci{0.0}}& & \textbf{97.1\ci{0.6}} & \textbf{24.6\ci{6.3}} & 
  & \textbf{84.3\ci{0.7}} & \textbf{311.1\ci{25.9}}& & \textbf{98.7\ci{0.3}} & \textbf{27.7\ci{6.4}} & 
  & \textbf{96.1\ci{0.5}} & \textbf{30.1\ci{5.3}}& & \textbf{82.1\ci{0.6}} & \textbf{113.1\ci{14.6}} &
  & \textbf{96.2\ci{0.6}} & \textbf{27.0\ci{4.9}}& & \textbf{97.9\ci{0.4}} & \textbf{15.6\ci{6.2}} \\\cmidrule(r){1-26}

  \ourrow \multicolumn{21}{l}{\textbf{(b) Ablation on Perceptual Network}} & \\ 
  \cdashline{1-26}\noalign{\vskip 0.6mm}
Sparse-3D-CNN & 
 & 100.0\ci{0.0} & 0.0\ci{0.0}& & 86.7\ci{2.0} & 143.5\ci{46.1} & 
  & 84.1\ci{1.5} & \textbf{277.8\ci{22.1}}& & 98.0\ci{.06} & 74.8\ci{7.9} & 
  & 88.8\ci{1.5} & 96.8\ci{11.6}& & 52.4\ci{1.5} & 365.9\ci{12.3} &
  & 80.1\ci{2.2} & 107.7\ci{15.8}& & 97.5\ci{0.4} & 18.9\ci{14.1} \\
3D-CNN & 
 & 99.9\ci{0.1} & 0.0\ci{0.0}& & \textbf{97.5\ci{0.5}} & \textbf{20.0\ci{6.6}} & 
  & 73.9\ci{2.1} & 379.0\ci{70.2}& & 96.1\ci{0.7} & 69.58\ci{5.8} & 
  & 92.7\ci{1.0} & 65.6\ci{9.5}& & 65.3\ci{0.9} & 275.4\ci{31.5} &
  & 86.0\ci{1.4} & 78.1\ci{19.2}& & \textbf{99.0\ci{0.3}} & 12.1\ci{11.6} \\
Sparse-2D-CNN & 
 & 99.6\ci{0.2} & 0.7\ci{1.4}& & 96.0\ci{1.0} & 26.17\ci{5.1} & 
  & 80.2\ci{1.1} & 363.1\ci{14.4}& & 92.7\ci{1.0} & 199.6\ci{120.2} & 
  & 87.9\ci{1.1} & 100.5\ci{20.3}& & 57.6\ci{0.9} & 360.3\ci{16.3} &
  & 89.1\ci{0.7} & 52.9\ci{4.8}& & 98.7\ci{0.6} & \textbf{4.55\ci{2.92}} \\
  
\ourshort & 
 & \textbf{100.0\ci{0.0}} & \textbf{0.0\ci{0.0}}& & 97.1\ci{0.6} & 24.6\ci{6.3} & 
  & \textbf{84.3\ci{0.7}} & 311.1\ci{25.9}& & \textbf{98.7\ci{0.3}} & \textbf{27.7\ci{6.4}} & 
  & \textbf{96.1\ci{0.5}} & \textbf{30.1\ci{5.3}}& & \textbf{82.1\ci{0.6}} & \textbf{113.1\ci{14.6}} &
  & \textbf{96.2\ci{0.6}} & \textbf{27.0\ci{4.9}}& & 97.9\ci{0.4} & 15.6\ci{6.2} \\

 \midrule 
  \ourrow \multicolumn{21}{l}{\textbf{(c) Ablation on Perceptual Interface}} & \\
  \cdashline{1-26}\noalign{\vskip 0.6mm}
Only-Height-Map & 
  & 100.0\ci{0.0} & 0.0\ci{0.0}& & 5.3\ci{2.0} & 1995.3\ci{68.3} & 
  & 10.5\ci{1.5} & 577.4\ci{18.1}& & 10.2\ci{1.3} & 717.5\ci{33.8} & 
  & 96.0\ci{0.7} & 34.3\ci{2.8}& & \textbf{86.2\ci{0.6}} & \textbf{101.6\ci{13.8}} &
  & \textbf{98.3\ci{0.2}} & \textbf{11.6\ci{6.2}}& & 98.5\ci{0.3} & 11.2\ci{6.7} \\
Only-Voxel-Grid & 
  & 100.0\ci{0.0} & 0.0\ci{0.0}& & 96.9\ci{0.4} & \textbf{22.4\ci{4.2}} & 
  & 75.9\ci{1.5} & 506.0\ci{20.6}& & 96.0\ci{0.3} & 281.4\ci{29.0} & 
  & 94.2\ci{0.8} & 51.0\ci{10.2}& & 72.3\ci{0.6} & 201.8\ci{14.9} &
  & 89.3\ci{1.3} & 46.9\ci{10.5}& & \textbf{98.8\ci{0.2}} & \textbf{7.0\ci{3.9}} \\
  
\ourshort & 
 & \textbf{100.0\ci{0.0}} & \textbf{0.0\ci{0.0}}& &\textbf{ 97.1\ci{0.6}} & 24.6\ci{6.3} & 
  & \textbf{84.3\ci{0.7}} & \textbf{311.1\ci{25.9}}& &\textbf{ 98.7\ci{0.3}} & \textbf{27.7\ci{6.4}} & 
  & \textbf{96.1\ci{0.5}} & \textbf{30.1\ci{5.3}}& & 82.1\ci{0.6} & 113.1\ci{14.6} &
  & 96.2\ci{0.6} & 27.0\ci{4.9}& & 97.9\ci{0.4} & 15.6\ci{6.2} \\ \cmidrule(r){1-26}

  \ourrow \multicolumn{21}{l}{\textbf{(d) Ablation on Voxel Resolution}} & \\ 
  \cdashline{1-26}\noalign{\vskip 0.6mm}
10CM & 
  & 98.8\ci{0.2} & 2.1\ci{1.6}& & \textbf{97.3\ci{0.9}} & \textbf{24.2\ci{11.0}} & 
  & 77.5\ci{3.4} & 368.0\ci{36.3}& & 97.5\ci{0.4} & 260.4\ci{38.8} & 
  & 75.5\ci{0.5} & 63.0\ci{4.9}& & 65.2\ci{5.5} & 256.3\ci{50.0} &
  & 94.1\ci{1.1} & 38.6\ci{6.7}& & 97.5\ci{0.4} & \textbf{13.5\ci{2.0}} \\
  
2.5CM & 
  & 99.9\ci{0.1} & 2.1\ci{1.6}& & 13.3\ci{2.4} & 1442.4\ci{119.6} & 
  & 59.0\ci{1.7} & 642.7\ci{12.4}& & 64.8\ci{1.1} & 591.0\ci{22.5} & 
  & 67.2\ci{2.7} & 268.9\ci{39.3}& & 54.1\ci{1.7} & 400.2\ci{19.5} &
  & 86.3\ci{1.2} & 74.8\ci{12.8}& & 96.6\ci{0.4} & 15.2\ci{6.1} \\

\ourshort (5CM) & 
 & \textbf{100.0\ci{0.0}} & \textbf{0.0\ci{0.0}}& & 97.1\ci{0.6} & 24.6\ci{6.3} & 
  & \textbf{84.3\ci{0.7}} & \textbf{311.1\ci{25.9}}& & \textbf{98.7\ci{0.3}} & \textbf{27.7\ci{6.4}} & 
  & \textbf{96.1\ci{0.5}} & \textbf{30.1\ci{5.3}}& & \textbf{82.1\ci{0.6}} & \textbf{113.1\ci{14.6}} &
  & \textbf{96.2\ci{0.6}} & \textbf{27.0\ci{4.9}}& & \textbf{97.9\ci{0.4}} & 15.6\ci{6.2} \\
\bottomrule

\end{tabular}}
\label{table:main_results}
\vspace{-0.4cm}
\end{table*}

\subsection{Experimental Configuration}
\begin{figure*}[!ht]
\vspace{8pt}
    \centering
    \includegraphics[width=1.0\linewidth]{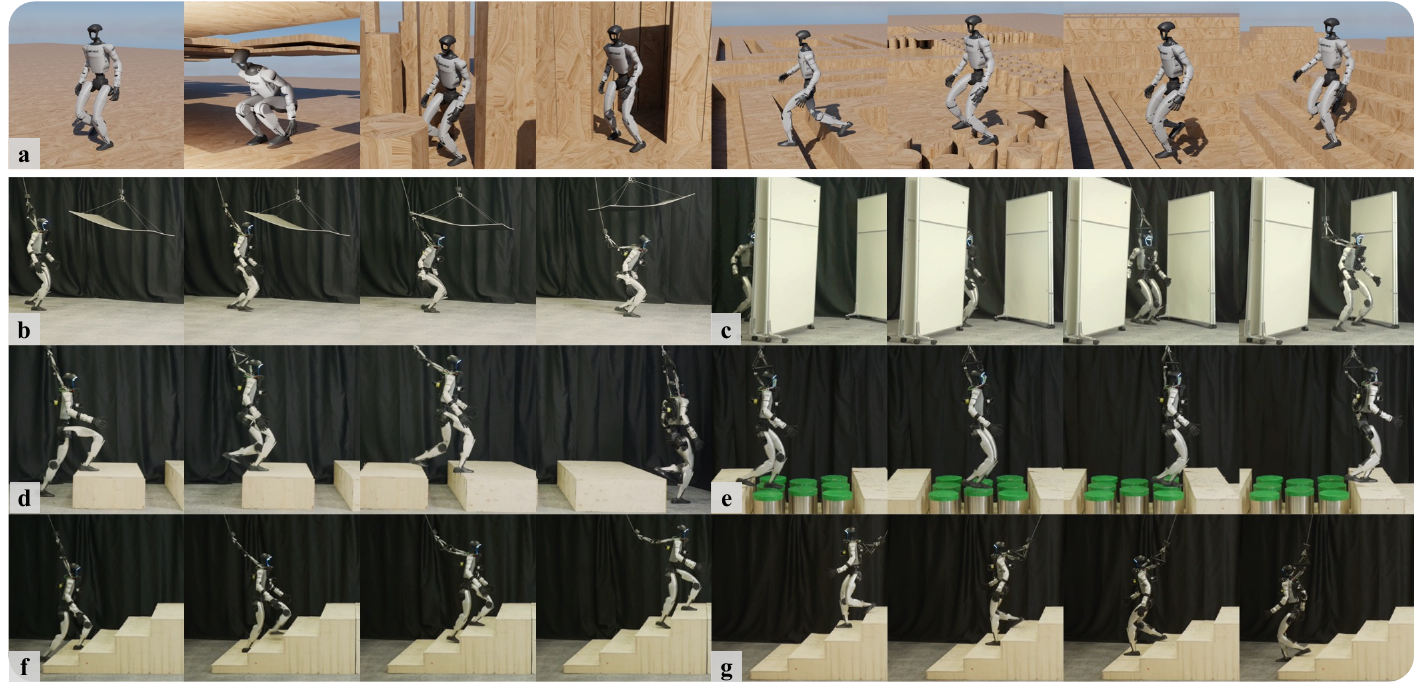}
    \vspace{-15pt}
    \caption{Humanoid robot traverses diverse 3D constrained terrains in both simulation and the real world.
(a)Traversal across the eight simulated training terrain types.
(b)Ducking under suspended ceiling obstacles.
(c)Local navigation through lateral clutters.
(d)Stepping onto a $30cm$-high platform and crossing a $40cm$ gap.
(e)Traversing pile-like stepping-stone terrain.
(f)(g)Ascending and descending $20cm$ stairs. All deployments are based on the \textbf{same policy}.}
    \label{fig:cross}
\end{figure*}
\label{sec:method_terrain}

We conduct both simulation training and real-world deployment on the 29-DoF Unitree G1 humanoid. Simulation is performed using NVIDIA IsaacSim~\cite{NVIDIA_Isaac_Sim}. To ensure voxel grids accurately capture full-space terrain geometry, we mount two Hesai JT128 LiDARs on the robot—one on the front chest and one on the back—each with a $95^\circ \times 360^\circ$ field of view. This dual-sensor configuration is identically replicated in simulation to ensure consistent perception across domains. Policy training is distributed across eight NVIDIA RTX 4090 GPUs (45GB memory each). During deployment, both the learned policy and voxel grid processing run entirely onboard the G1 using an NVIDIA Orin NX. For target-relative localization, we use a Livox Mid-360 LiDAR mounted on the robot’s head and process its data using FastLIO2~\cite{xu2022fast,xu2021fast}. This LiDAR also provides input for elevation map generation in baseline comparisons.

\subsection{Simulation Experiments}
\subsubsection{Metrics}
We evaluate ablated methods in IsaacSim~\cite{NVIDIA_Isaac_Sim} on the most challenging terrain settings ($\mathbf{p}_\tau^{\max}$; \cref{sec:method_terrain}), and the policy performance is measured by two distinct metrics: 
\begin{itemize}
    \item \textbf{Success rate \(E_{\mathrm{succ}}\)}: fraction of episodes that reach the target within a 10s horizon without falling or incurring any severe collisions with the obstacles.
    \item \textbf{Collision momentum \(E_{\mathrm{collision}}\)}: cumulative momentum transferred through unnecessary contacts (all robot–environment contacts excluding nominal foot contacts), reflecting the policy’s ability to avoid collisions.
\end{itemize} 
We train every policy for \(4{,}000\) iterations, then run \(5\) independent evaluations (each run evaluates over \(1{,}000\) complete episodes), reporting mean \(\pm\) standard deviation; policies with higher \(E_{\mathrm{succ}}\) and lower \(E_{\mathrm{col}}\) are better.

\subsubsection{Baselines}
To assess the effectiveness of core components in Gallant, we compare against the following ablations:
\begin{itemize}
    \item \textbf{Self-scan.} We disable simulated LiDAR returns from dynamic geometry (e.g. the robot’s own links), but only scans static terrain. This is compared to \ourshort, which models scans over both static terrain and moving links.
    \item \textbf{Perceptual network.} We replace the \(z\)-as-channel 2D CNN with alternatives: standard 3D CNN, sparse 2D CNN, and sparse 3D CNN (commonly used in LiDAR perception~\cite{frey2022locomotion,chen2023voxelnext}). Sparse variants are based on \cite{spconv2022}.
    \item \textbf{Perceptual representation.} Gallant feeds a voxel grid to the actor and a voxel grid plus a height map to the critic. We test two baselines to isolate this: (i) only height map for actor and critic; (ii) only voxel grid for actor and critic.
    \item \textbf{Voxel resolution.} We sweep the voxel size around the default \(5\)\,cm (i.e., \(2.5\)\,cm and \(10\)\,cm) to examine the trade-off between field of view coverage and geometric fidelity.
\end{itemize}

\subsubsection{Result}

Across eight representative terrains, Gallant attains superior success rates relative to the baselines (see \cref{table:main_results}). Ablation-specific analyses are summarized as follow: 

\textbf{LiDAR return from dynamic objects is necessary.}
With all other settings fixed, Gallant achieves much higher success rates than the variant that ignores dynamic objects (w/o-Self-Scan) across all tasks. Using Ceiling as an example in \cref{fig:simresult} (a), when the robot ducks under the ceiling, the voxel grids with dynamics (\cref{fig:simresult} (b)) correctly include the robot’s legs, which occupy voxels and induce occlusion “holes” along LiDAR rays to the distant floor. In contrast, excluding dynamics (\cref{fig:simresult} (c)) yields an artificially flat floor. Because real LiDAR returns from all visible objects, omitting dynamics makes the voxel grid out-of-distribution (OOD) in postures where the body is not fully upright (e.g., Ceiling, Platform), causing a pronounced drop in success. Hence, simulating dynamic objects in the LiDAR pipeline is critical to final performance.
\begin{figure*}[!ht]
    \centering
    \includegraphics[width=1.0\linewidth]{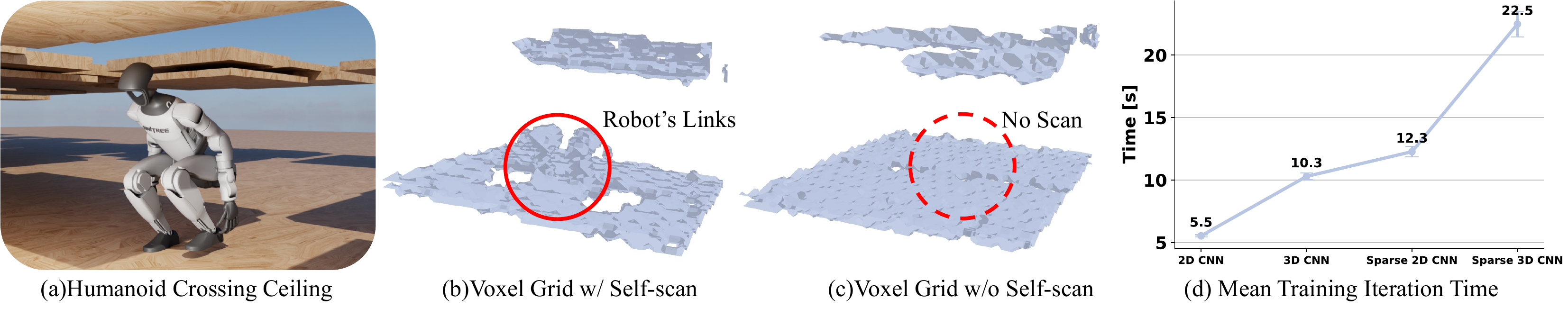}
    \caption{
\textbf{Visualization of simulation ablation analyses.} 
(a) The humanoid crouches to traverse under a low ceiling; 
(b) Voxel grid from LiDAR simulation that includes dynamic objects captures the robot’s own links; 
(c) LiDAR simulation restricted to static objects excludes robot links from the voxel grid; 
(d) Mean training iteration time for Gallant with different CNN-based perception modules.
}
\vspace{-10pt}
    \label{fig:simresult}
    
\end{figure*}
\textbf{$z$-grouped 2D CNN is the most suitable choice.}
Although one variant marginally exceeds Gallant on a few terrains (e.g., a 3D CNN on Ceiling), the gains are small and are outweighed by lower success rates on most tasks. Our voxel input is a compact, egocentric grid of \(32\times32\times40\), which changes with the torso frame. As shown in \cref{fig:simresult} (d), sparse convolutions offer little advantage: occupancy is relatively dense in the \(x\!-\!y\) plane, so few computations are actually skipped, while the rulebook overhead of sparse kernels becomes a dominant cost at this scale. On the other hand, full 3D CNNs introduce substantially more parameters and memory traffic, making optimization harder and less data-efficient when sparsity is concentrated primarily along \(z\). Treating \(z\) as channels with a lightweight 2D CNN preserves vertical structure through channel mixing, exploits highly optimized dense 2D operators, and provides the right inductive bias for an egocentric raster that is approximately translation-equivariant in \(x\!-\!y\) yet rotates with the body. In practice, this z-grouped 2D design delivers equal or better accuracy with markedly lower compute, making it the most suitable choice for our task.

\textbf{Combination of Voxel Grid and Height Map is better.}
As discussed in \cref{sec:intro}, using only a height map as the perceptual representation for policy cannot represent multi-layer structure; consequently, Only-Height-Map fails on terrains such as Ceiling. Nevertheless, in simulation (where the height map incurs no latency), height-map–based methods perform strongly on ground obstacles, indicating that the height map provides a useful, positively informative signal for training. For sim-to-real robustness, Gallant therefore omits the height map from the actor inputs, but includes it as part of the critic observation (privileged information). This asymmetric design leverages the height map to shape values and improve credit assignment during training while keeping the deployed policy free of latency-sensitive channels. This Gallant configuration achieves higher success rates than Only-Voxel-Grid (critic without height map) across all tasks, validating the proposed design.

\textbf{5cm is a suitable resolution for Gallant.}
PPO training benefits from large batches collected over many parallel environments. Under a fixed VRAM budget, we therefore adjust the voxel-grid resolution to trade spatial precision for egocentric FoV. Empirically, the \(10\,\mathrm{cm}\) grid underperforms Gallant’s \(5\,\mathrm{cm}\) setting: while it enlarges the FOV, its coarse quantization impairs fine contact- and clearance-sensitive interactions. Conversely, the \(2.5\,\mathrm{cm}\) grid yields an even lower success rate: despite its higher precision, the reduced FOV hampers perception of long vertical extents, making terrains that require sensing far below or above the robot (e.g., Ceiling, Downstair) notably harder. Overall, the \(5\,\mathrm{cm}\) resolution strikes an effective balance between coverage and detail under resource constraints.

\subsection{Real-world Experiments}
\subsubsection{Deployment}
We directly deploy the Gallant-trained policy onto the real Unitree G1 humanoid without any fine-tuning. The control loop runs at $50Hz$, consistent with simulation. To ensure reliable voxel input, raw point clouds from dual LiDARs are processed onboard using OctoMap~\cite{hornung13auro}, generating a binary occupancy grid at $10Hz$. Importantly, OctoMap serves as a lightweight preprocessing step—not a full reconstruction pipeline like elevation maps—and thus incurs minimal latency or computational load.

We evaluate the same policy across a variety of real-world scenarios, including flat terrain, random-height ceilings, lateral clutters (e.g., doors), high platforms with gaps, stepping stones, and staircases. Despite the diverse and complex constraints, the robot consistently traverses these terrains with high success rates (see~\cref{fig:realexp}). Qualitative results are shown in~\cref{fig:cross}. The policy exhibits versatile capabilities: it crouches under ceilings of varying heights, plans lateral motions to pass through narrow doorways, steps robustly onto high platforms, crosses gaps between them, and carefully places its feet to negotiate stepping-stone–like terrains. On stairs, it demonstrates stable multi-step climbing and descent without loss of balance. These results highlight Gallant’s ability to encode spatial constraints from perception and translate them into robust, real-time whole-body behaviors. Furthermore, all behaviors arise from a single policy without any terrain-specific tuning, highlighting Gallant’s generality and real-world transferability.

\subsubsection{Ablation}
\label{sec:real_aba}
\begin{figure*}[!ht]
    \centering
    \includegraphics[width=1.0\linewidth]{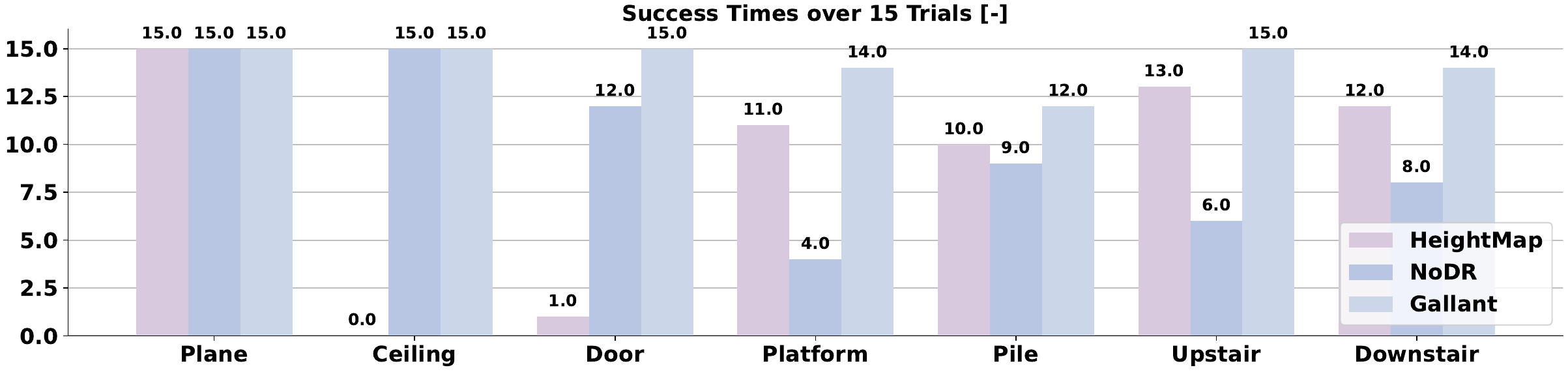}
    \caption{\textbf{Real-world traversal success times over 15 trials.} \textit{Height Map} uses elevation maps as perceptual representation; \textit{NoDR} is Gallant without LiDAR domain randomization; Gallant denotes the full proposed pipeline. All methods are tested for 15 trials per terrain.}
    \label{fig:realexp}
    \vspace{-10pt}
\end{figure*}
To evaluate sim-to-real performance, we deploy three policies on the 29-DoF Unitree G1 and compare success rates across terrains: (i) HeightMap, which replaces the voxel grid with an elevation map estimated from Livox Mid360; (ii) NoDR, trained without the LiDAR domain randomization described in Sec.~\ref{sec:method_lidar}, but otherwise identical to Gallant; and (iii) Gallant, our full pipeline. Each policy is tested over 15 trials per terrain, with results shown in~\cref{fig:realexp}.

Gallant consistently outperforms both baselines across all real-world terrains. The HeightMap baseline fails on overheading (e.g., Ceiling) and lateral (e.g., Door) obstacles due to its limited 2.5D representation, and performs worse than Gallant even on ground-level terrains. Unlike in simulation, where HeightMap occasionally excels on Pile or Stairs, its real-world performance is hindered by noisy elevation reconstruction. Moreover, our policy allows torso pitch/roll for more expressive motion, but this introduces LiDAR jitter at the mounting point, further degrading elevation map quality—reinforcing the benefit of voxel grids. The NoDR variant performs reasonably well on Ceiling and Door, suggesting low sensitivity to sensing latency in these cases. However, its performance drops significantly on ground-level terrains. Without modeling LiDAR delay and noise in training, the robot misjudges its position relative to obstacles, often reacting too late. This emphasizes the critical role of domain randomization in bridging the sim-to-real gap.


\subsection{Further Analyses}
\begin{figure}[!ht]
    \centering
    \includegraphics[width=0.95\linewidth]{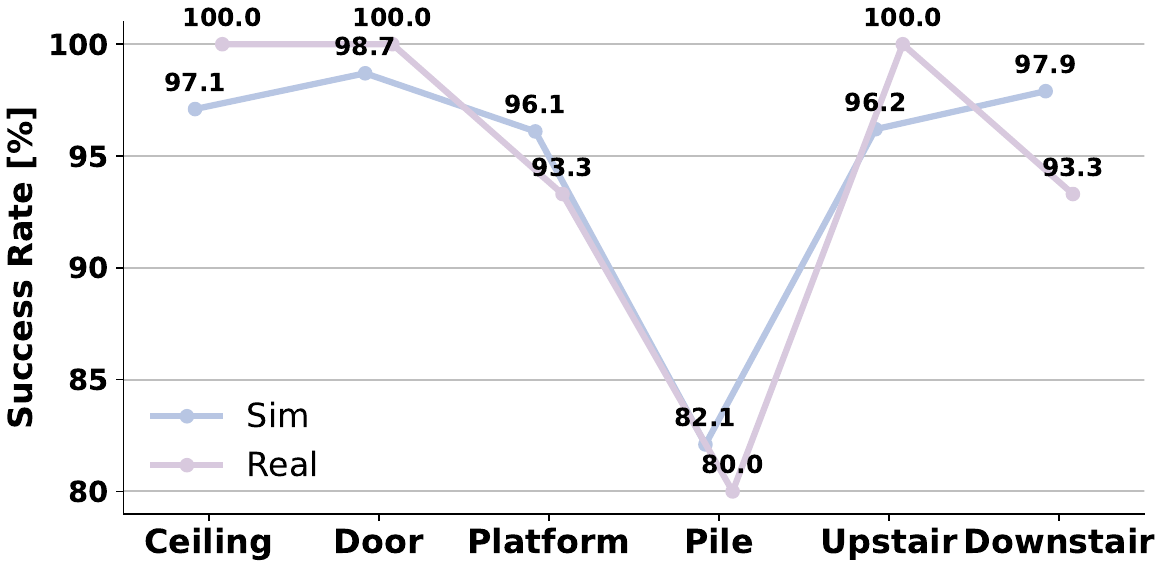}
    \caption{Gallant success rate in simulation and real world.}
    \label{fig:succ}
    \vspace{-5pt}
\end{figure}
We analyze Gallant’s success rates across terrains evaluated in both simulation and the real world (\cref{fig:succ}). A clear correlation emerges: terrains with higher success in simulation also perform well on hardware, validating the use of large-scale simulated evaluation as a reliable predictor of real-world performance. With the introduction of voxel grids, scenarios like overheading (e.g., Ceiling) and lateral (e.g., Door) constraints—previously difficult for height map-based methods—become the easiest considering the high success rate, demonstrating voxel grids as a simple yet effective representation for full-space perception.

Gallant’s main limitation appears on the Pile terrain, where accurate foothold selection is critical. Success rates plateau around 80\%, and simulation with zero LiDAR latency improves this to over 90\%, indicating that real-world sensor delay is a key bottleneck. On other terrains—especially Platforms and Stairs, previously considered unstable due to collision risk~\cite{long2025learning}—Gallant achieves high success by proactively adjusting foot trajectories.

\section{Conclusion}
\label{sec:conclusion}
We present Gallant, a full-stack pipeline for humanoid locomotion and local navigation in 3D-constrained environments. It leverages voxel grids as a lightweight, geometry-preserving perceptual representation, combined with realistic LiDAR simulation and a $z$-grouped 2D CNN for efficient processing. Simulation ablations show that Gallant’s key components are essential for training high–success-rate policies. In real-world tests, a single LiDAR policy covers the ground obstacles handled by elevation-map controllers while also tackling lateral and overhead structures, and on ground-only terrains it reaches near-100\% success with fewer collisions. All these results together establish Gallant as a solid pipeline for humanoid locomotion and local navigation across 3D-constrained terrains.

\textbf{Limitations.} Despite its success, Gallant does not yet achieve a 100\% success rate. The primary bottleneck lies in LiDAR latency: operating at 10 Hz, each scan incurs over 100 ms delay due to light reflection and communication overhead. This delay limits the robot’s ability to act preemptively. Future work will explore using Gallant as a geometry-aware teacher while investigating lower-latency sensors to enable a fully reactive policy that achieves near-perfect performance across all terrains.
\newpage
{
    \small
    \bibliographystyle{ieeenat_fullname}
    \bibliography{main}

\begin{thebibliography}{49}
\providecommand{\natexlab}[1]{#1}
\providecommand{\url}[1]{\texttt{#1}}
\expandafter\ifx\csname urlstyle\endcsname\relax
  \providecommand{\doi}[1]{doi: #1}\else
  \providecommand{\doi}{doi: \begingroup \urlstyle{rm}\Url}\fi

\bibitem[Agarwal et~al.(2023)Agarwal, Kumar, Malik, and Pathak]{agarwal2023legged}
Ananye Agarwal, Ashish Kumar, Jitendra Malik, and Deepak Pathak.
\newblock Legged locomotion in challenging terrains using egocentric vision.
\newblock In \emph{Conference on robot learning}, pages 403--415. PMLR, 2023.

\bibitem[Allshire et~al.(2025)Allshire, Choi, Zhang, McAllister, Zhang, Kim, Darrell, Abbeel, Malik, and Kanazawa]{videomimic}
Arthur Allshire, Hongsuk Choi, Junyi Zhang, David McAllister, Anthony Zhang, Chung~Min Kim, Trevor Darrell, Pieter Abbeel, Jitendra Malik, and Angjoo Kanazawa.
\newblock Visual imitation enables contextual humanoid control.
\newblock \emph{arXiv preprint arXiv:2505.03729}, 2025.

\bibitem[Ben et~al.(2025)Ben, Jia, Zeng, Dong, Lin, and Pang]{ben2025homie}
Qingwei Ben, Feiyu Jia, Jia Zeng, Junting Dong, Dahua Lin, and Jiangmiao Pang.
\newblock Homie: Humanoid loco-manipulation with isomorphic exoskeleton cockpit.
\newblock \emph{arXiv preprint arXiv:2502.13013}, 2025.

\bibitem[Cai et~al.(2025)Cai, Peng, Yang, Zhang, Wei, Wang, Chen, Wang, and Pang]{cai2025navdp}
Wenzhe Cai, Jiaqi Peng, Yuqiang Yang, Yujian Zhang, Meng Wei, Hanqing Wang, Yilun Chen, Tai Wang, and Jiangmiao Pang.
\newblock Navdp: Learning sim-to-real navigation diffusion policy with privileged information guidance.
\newblock \emph{arXiv preprint arXiv:2505.08712}, 2025.

\bibitem[Chen et~al.(2023)Chen, Liu, Zhang, Qi, and Jia]{chen2023voxelnext}
Yukang Chen, Jianhui Liu, Xiangyu Zhang, Xiaojuan Qi, and Jiaya Jia.
\newblock Voxelnext: Fully sparse voxelnet for 3d object detection and tracking.
\newblock In \emph{Proceedings of the IEEE/CVF conference on computer vision and pattern recognition}, pages 21674--21683, 2023.

\bibitem[Cheng et~al.(2024{\natexlab{a}})Cheng, Ji, Yang, Gongye, Zou, Kautz, B{\i}y{\i}k, Yin, Liu, and Wang]{cheng2024navila}
An-Chieh Cheng, Yandong Ji, Zhaojing Yang, Zaitian Gongye, Xueyan Zou, Jan Kautz, Erdem B{\i}y{\i}k, Hongxu Yin, Sifei Liu, and Xiaolong Wang.
\newblock Navila: Legged robot vision-language-action model for navigation.
\newblock \emph{arXiv preprint arXiv:2412.04453}, 2024{\natexlab{a}}.

\bibitem[Cheng et~al.(2024{\natexlab{b}})Cheng, Shi, Agarwal, and Pathak]{cheng2024extreme}
Xuxin Cheng, Kexin Shi, Ananye Agarwal, and Deepak Pathak.
\newblock Extreme parkour with legged robots.
\newblock In \emph{2024 IEEE International Conference on Robotics and Automation (ICRA)}, pages 11443--11450. IEEE, 2024{\natexlab{b}}.

\bibitem[Contributors(2022)]{spconv2022}
Spconv Contributors.
\newblock Spconv: Spatially sparse convolution library.
\newblock \url{https://github.com/traveller59/spconv}, 2022.

\bibitem[Cui et~al.(2025)Cui, Wang, Qin, Guo, Han, Zhao, Cao, Zhang, Zhong, Sun, et~al.]{cui2025humanoid}
Wei Cui, Haoyu Wang, Wenkang Qin, Yijie Guo, Gang Han, Wen Zhao, Jiahang Cao, Zhang Zhang, Jiaru Zhong, Jingkai Sun, et~al.
\newblock Humanoid occupancy: Enabling a generalized multimodal occupancy perception system on humanoid robots.
\newblock \emph{arXiv preprint arXiv:2507.20217}, 2025.

\bibitem[Fankhauser et~al.(2014)Fankhauser, Bloesch, Gehring, Hutter, and Siegwart]{Fankhauser2014RobotCentricElevationMapping}
P\'{e}ter Fankhauser, Michael Bloesch, Christian Gehring, Marco Hutter, and Roland Siegwart.
\newblock Robot-centric elevation mapping with uncertainty estimates.
\newblock In \emph{International Conference on Climbing and Walking Robots (CLAWAR)}, 2014.

\bibitem[Fankhauser et~al.(2018)Fankhauser, Bloesch, and Hutter]{Fankhauser2018ProbabilisticTerrainMapping}
P{\'{e}}ter Fankhauser, Michael Bloesch, and Marco Hutter.
\newblock Probabilistic terrain mapping for mobile robots with uncertain localization.
\newblock \emph{IEEE Robotics and Automation Letters (RA-L)}, 3\penalty0 (4):\penalty0 3019--3026, 2018.

\bibitem[Frey et~al.(2022)Frey, Hoeller, Khattak, and Hutter]{frey2022locomotion}
Jonas Frey, David Hoeller, Shehryar Khattak, and Marco Hutter.
\newblock Locomotion policy guided traversability learning using volumetric representations of complex environments.
\newblock In \emph{2022 IEEE/RSJ International Conference on Intelligent Robots and Systems (IROS)}, pages 5722--5729. IEEE, 2022.

\bibitem[He et~al.(2025)He, Zhang, Jenelten, Grandia, B{\"a}cher, and Hutter]{he2025attention}
Junzhe He, Chong Zhang, Fabian Jenelten, Ruben Grandia, Moritz B{\"a}cher, and Marco Hutter.
\newblock Attention-based map encoding for learning generalized legged locomotion.
\newblock \emph{Science Robotics}, 10\penalty0 (105):\penalty0 eadv3604, 2025.

\bibitem[He et~al.(2024)He, Zhang, Xiao, He, Liu, and Shi]{he2024agile}
Tairan He, Chong Zhang, Wenli Xiao, Guanqi He, Changliu Liu, and Guanya Shi.
\newblock Agile but safe: Learning collision-free high-speed legged locomotion.
\newblock \emph{arXiv preprint arXiv:2401.17583}, 2024.

\bibitem[Hoeller et~al.(2024)Hoeller, Rudin, Sako, and Hutter]{hoeller2024anymal}
David Hoeller, Nikita Rudin, Dhionis Sako, and Marco Hutter.
\newblock Anymal parkour: Learning agile navigation for quadrupedal robots.
\newblock \emph{Science Robotics}, 9\penalty0 (88):\penalty0 eadi7566, 2024.

\bibitem[Hornung et~al.(2013)Hornung, Wurm, Bennewitz, Stachniss, and Burgard]{hornung13auro}
Armin Hornung, Kai~M. Wurm, Maren Bennewitz, Cyrill Stachniss, and Wolfram Burgard.
\newblock {OctoMap}: An efficient probabilistic {3D} mapping framework based on octrees.
\newblock \emph{Autonomous Robots}, 2013.
\newblock Software available at \url{https://octomap.github.io}.

\bibitem[Intelligence et~al.()Intelligence, Black, Brown, Darpinian, Dhabalia, Driess, Esmail, Equi, Finn, Fusai, et~al.]{intelligence2504pi0}
Physical Intelligence, Kevin Black, Noah Brown, James Darpinian, Karan Dhabalia, Danny Driess, Adnan Esmail, Michael Equi, Chelsea Finn, Niccolo Fusai, et~al.
\newblock $\pi$0. 5: a vision-language-action model with open-world generalization, 2025.
\newblock \emph{URL https://arxiv. org/abs/2504.16054}, 1\penalty0 (2):\penalty0 3.

\bibitem[Li et~al.(2025)Li, Luo, Wu, and Zhu]{li2025move}
Songbo Li, Shixin Luo, Jun Wu, and Qiuguo Zhu.
\newblock Move: Multi-skill omnidirectional legged locomotion with limited view in 3d environments.
\newblock In \emph{2025 IEEE International Conference on Robotics and Automation (ICRA)}, pages 7647--7653. IEEE, 2025.

\bibitem[Liao et~al.(2025)Liao, Truong, Huang, Tevet, Sreenath, and Liu]{liao2025beyondmimic}
Qiayuan Liao, Takara~E Truong, Xiaoyu Huang, Guy Tevet, Koushil Sreenath, and C~Karen Liu.
\newblock Beyondmimic: From motion tracking to versatile humanoid control via guided diffusion.
\newblock \emph{arXiv preprint arXiv:2508.08241}, 2025.

\bibitem[Liu et~al.(2024)Liu, Chen, Cheng, Ji, Qiu, Yang, and Wang]{liu2024visual}
Minghuan Liu, Zixuan Chen, Xuxin Cheng, Yandong Ji, Ri-Zhao Qiu, Ruihan Yang, and Xiaolong Wang.
\newblock Visual whole-body control for legged loco-manipulation.
\newblock \emph{arXiv preprint arXiv:2403.16967}, 2024.

\bibitem[Long et~al.(2025)Long, Ren, Shi, Wang, Huang, Luo, and Pang]{long2025learning}
Junfeng Long, Junli Ren, Moji Shi, Zirui Wang, Tao Huang, Ping Luo, and Jiangmiao Pang.
\newblock Learning humanoid locomotion with perceptive internal model.
\newblock In \emph{2025 IEEE International Conference on Robotics and Automation (ICRA)}, pages 9997--10003. IEEE, 2025.

\bibitem[Loquercio et~al.(2023)Loquercio, Kumar, and Malik]{loquercio2023learning}
Antonio Loquercio, Ashish Kumar, and Jitendra Malik.
\newblock Learning visual locomotion with cross-modal supervision.
\newblock In \emph{IEEE International Conference on Robotics and Automation (ICRA)}, pages 7295--7302. IEEE, 2023.

\bibitem[Luo et~al.(2025)Luo, Yuan, Wang, Li, Chen, Casta{\~n}eda, Cao, Li, Minor, Ben, et~al.]{luo2025sonic}
Zhengyi Luo, Ye Yuan, Tingwu Wang, Chenran Li, Sirui Chen, Fernando Casta{\~n}eda, Zi-Ang Cao, Jiefeng Li, David Minor, Qingwei Ben, et~al.
\newblock Sonic: Supersizing motion tracking for natural humanoid whole-body control.
\newblock \emph{arXiv preprint arXiv:2511.07820}, 2025.

\bibitem[Macklin(2022)]{warp2022}
Miles Macklin.
\newblock Warp: A high-performance python framework for gpu simulation and graphics.
\newblock \url{https://github.com/nvidia/warp}, 2022.
\newblock NVIDIA GPU Technology Conference (GTC).

\bibitem[Maturana and Scherer(2015)]{maturana2015voxnet}
Daniel Maturana and Sebastian Scherer.
\newblock Voxnet: A 3d convolutional neural network for real-time object recognition.
\newblock In \emph{2015 IEEE/RSJ international conference on intelligent robots and systems (IROS)}, pages 922--928. Ieee, 2015.

\bibitem[Miki et~al.(2022)Miki, Wellhausen, Grandia, Jenelten, Homberger, and Hutter]{miki2022elevation}
Takahiro Miki, Lorenz Wellhausen, Ruben Grandia, Fabian Jenelten, Timon Homberger, and Marco Hutter.
\newblock Elevation mapping for locomotion and navigation using gpu.
\newblock In \emph{2022 IEEE/RSJ International Conference on Intelligent Robots and Systems (IROS)}, pages 2273--2280. IEEE, 2022.

\bibitem[Niijima et~al.(2025)Niijima, Tsuzaki, Takasugi, and Kinoshita]{niijima2025real}
Shun Niijima, Ryoichi Tsuzaki, Noriaki Takasugi, and Masaya Kinoshita.
\newblock Real-time multi-plane segmentation based on gpu accelerated high-resolution 3d voxel mapping for legged robot locomotion.
\newblock \emph{arXiv preprint arXiv:2510.01592}, 2025.

\bibitem[{NVIDIA}()]{NVIDIA_Isaac_Sim}
{NVIDIA}.
\newblock {Isaac Sim}.

\bibitem[Qiu et~al.(2025)Qiu, Song, Peng, Suryadevara, Yang, Liu, Ji, Jia, Yang, Zou, et~al.]{qiu2025wildlma}
Ri-Zhao Qiu, Yuchen Song, Xuanbin Peng, Sai~Aneesh Suryadevara, Ge Yang, Minghuan Liu, Mazeyu Ji, Chengzhe Jia, Ruihan Yang, Xueyan Zou, et~al.
\newblock Wildlma: Long horizon loco-manipulation in the wild.
\newblock In \emph{2025 IEEE International Conference on Robotics and Automation (ICRA)}, pages 10011--10019. IEEE, 2025.

\bibitem[Ren et~al.(2025)Ren, Huang, Wang, Wang, Ben, Long, Yang, Pang, and Luo]{ren2025vb}
Junli Ren, Tao Huang, Huayi Wang, Zirui Wang, Qingwei Ben, Junfeng Long, Yanchao Yang, Jiangmiao Pang, and Ping Luo.
\newblock Vb-com: Learning vision-blind composite humanoid locomotion against deficient perception.
\newblock \emph{arXiv preprint arXiv:2502.14814}, 2025.

\bibitem[Rudin et~al.(2022)Rudin, Hoeller, Bjelonic, and Hutter]{rudin2022advanced}
Nikita Rudin, David Hoeller, Marko Bjelonic, and Marco Hutter.
\newblock Advanced skills by learning locomotion and local navigation end-to-end.
\newblock In \emph{2022 IEEE/RSJ International Conference on Intelligent Robots and Systems (IROS)}, pages 2497--2503. IEEE, 2022.

\bibitem[Schulman et~al.(2017)Schulman, Wolski, Dhariwal, Radford, and Klimov]{schulman2017proximal}
John Schulman, Filip Wolski, Prafulla Dhariwal, Alec Radford, and Oleg Klimov.
\newblock Proximal policy optimization algorithms.
\newblock \emph{arXiv preprint arXiv:1707.06347}, 2017.

\bibitem[Shi et~al.(2025)Shi, Wang, Guo, Duan, Wang, Chen, Yang, Wang, and Wang]{shi2025oneocc}
Hao Shi, Ze Wang, Shangwei Guo, Mengfei Duan, Song Wang, Teng Chen, Kailun Yang, Lin Wang, and Kaiwei Wang.
\newblock Oneocc: Semantic occupancy prediction for legged robots with a single panoramic camera.
\newblock \emph{arXiv preprint arXiv:2511.03571}, 2025.

\bibitem[Sun et~al.(2025{\natexlab{a}})Sun, Han, Sun, Zhao, Cao, Wang, Guo, and Zhang]{sun2025dpl}
Jingkai Sun, Gang Han, Pihai Sun, Wen Zhao, Jiahang Cao, Jiaxu Wang, Yijie Guo, and Qiang Zhang.
\newblock Dpl: Depth-only perceptive humanoid locomotion via realistic depth synthesis and cross-attention terrain reconstruction.
\newblock \emph{arXiv preprint arXiv:2510.07152}, 2025{\natexlab{a}}.

\bibitem[Sun et~al.(2025{\natexlab{b}})Sun, Cao, Chen, Su, Liu, Xie, and Liu]{sun2025learning}
Wandong Sun, Baoshi Cao, Long Chen, Yongbo Su, Yang Liu, Zongwu Xie, and Hong Liu.
\newblock Learning perceptive humanoid locomotion over challenging terrain.
\newblock \emph{arXiv preprint arXiv:2503.00692}, 2025{\natexlab{b}}.

\bibitem[Team(2025)]{generalist2025gen0}
Generalist~AI Team.
\newblock Gen-0: Embodied foundation models that scale with physical interaction.
\newblock \emph{Generalist AI Blog}, 2025.
\newblock https://generalistai.com/blog/preview-uqlxvb-bb.html.

\bibitem[Tran et~al.(2015)Tran, Bourdev, Fergus, Torresani, and Paluri]{tran2015learning}
Du Tran, Lubomir Bourdev, Rob Fergus, Lorenzo Torresani, and Manohar Paluri.
\newblock Learning spatiotemporal features with 3d convolutional networks.
\newblock In \emph{Proceedings of the IEEE international conference on computer vision}, pages 4489--4497, 2015.

\bibitem[Wang et~al.(2025{\natexlab{a}})Wang, Wang, Ren, Ben, Huang, Zhang, and Pang]{wang2025beamdojo}
Huayi Wang, Zirui Wang, Junli Ren, Qingwei Ben, Tao Huang, Weinan Zhang, and Jiangmiao Pang.
\newblock Beamdojo: Learning agile humanoid locomotion on sparse footholds.
\newblock \emph{arXiv preprint arXiv:2502.10363}, 2025{\natexlab{a}}.

\bibitem[Wang et~al.(2025{\natexlab{b}})Wang, Ma, Jia, Yang, Zhou, Ouyang, Zhang, and Liang]{wang2025omni}
Zifan Wang, Teli Ma, Yufei Jia, Xun Yang, Jiaming Zhou, Wenlong Ouyang, Qiang Zhang, and Junwei Liang.
\newblock Omni-perception: Omnidirectional collision avoidance for legged locomotion in dynamic environments.
\newblock \emph{arXiv preprint arXiv:2505.19214}, 2025{\natexlab{b}}.

\bibitem[Wei et~al.(2025)Wei, Wan, Yu, Wang, Yang, Mao, Zhu, Cai, Wang, Chen, et~al.]{wei2025streamvln}
Meng Wei, Chenyang Wan, Xiqian Yu, Tai Wang, Yuqiang Yang, Xiaohan Mao, Chenming Zhu, Wenzhe Cai, Hanqing Wang, Yilun Chen, et~al.
\newblock Streamvln: Streaming vision-and-language navigation via slowfast context modeling.
\newblock \emph{arXiv preprint arXiv:2507.05240}, 2025.

\bibitem[Xu et~al.(2025)Xu, Weng, Lu, Gao, and Xu]{xu2025facet}
Botian Xu, Haoyang Weng, Qingzhou Lu, Yang Gao, and Huazhe Xu.
\newblock Facet: Force-adaptive control via impedance reference tracking for legged robots.
\newblock \emph{arXiv preprint arXiv:2505.06883}, 2025.

\bibitem[Xu and Zhang(2021)]{xu2021fast}
Wei Xu and Fu Zhang.
\newblock Fast-lio: A fast, robust lidar-inertial odometry package by tightly-coupled iterated kalman filter.
\newblock \emph{IEEE Robotics and Automation Letters}, 6\penalty0 (2):\penalty0 3317--3324, 2021.

\bibitem[Xu et~al.(2022)Xu, Cai, He, Lin, and Zhang]{xu2022fast}
Wei Xu, Yixi Cai, Dongjiao He, Jiarong Lin, and Fu Zhang.
\newblock Fast-lio2: Fast direct lidar-inertial odometry.
\newblock \emph{IEEE Transactions on Robotics}, 38\penalty0 (4):\penalty0 2053--2073, 2022.

\bibitem[Yang et~al.(2021)Yang, Zhang, Hansen, Xu, and Wang]{yang2021learning}
Ruihan Yang, Minghao Zhang, Nicklas Hansen, Huazhe Xu, and Xiaolong Wang.
\newblock Learning vision-guided quadrupedal locomotion end-to-end with cross-modal transformers.
\newblock \emph{arXiv preprint arXiv:2107.03996}, 2021.

\bibitem[Yokoyama et~al.(2023)Yokoyama, Clegg, Truong, Undersander, Yang, Arnaud, Ha, Batra, and Rai]{yokoyama2023asc}
Naoki Yokoyama, Alex Clegg, Joanne Truong, Eric Undersander, Tsung-Yen Yang, Sergio Arnaud, Sehoon Ha, Dhruv Batra, and Akshara Rai.
\newblock Asc: Adaptive skill coordination for robotic mobile manipulation.
\newblock \emph{IEEE Robotics and Automation Letters}, 9\penalty0 (1):\penalty0 779--786, 2023.

\bibitem[Zhang et~al.(2024)Zhang, Jin, Frey, Rudin, Mattamala, Cadena, and Hutter]{zhang2024resilient}
Chong Zhang, Jin Jin, Jonas Frey, Nikita Rudin, Mat{\'\i}as Mattamala, Cesar Cadena, and Marco Hutter.
\newblock Resilient legged local navigation: Learning to traverse with compromised perception end-to-end.
\newblock In \emph{2024 IEEE International Conference on Robotics and Automation (ICRA)}, pages 34--41. IEEE, 2024.

\bibitem[Zhang et~al.(2025)Zhang, Zhang, Cui, Sun, Cao, Guo, Han, Zhao, Wang, Sun, et~al.]{zhang2025humanoidpano}
Qiang Zhang, Zhang Zhang, Wei Cui, Jingkai Sun, Jiahang Cao, Yijie Guo, Gang Han, Wen Zhao, Jiaxu Wang, Chenghao Sun, et~al.
\newblock Humanoidpano: Hybrid spherical panoramic-lidar cross-modal perception for humanoid robots.
\newblock \emph{arXiv preprint arXiv:2503.09010}, 2025.

\bibitem[Zhuang et~al.(2023)Zhuang, Fu, Wang, Atkeson, Schwertfeger, Finn, and Zhao]{zhuang2023robot}
Ziwen Zhuang, Zipeng Fu, Jianren Wang, Christopher Atkeson, Soeren Schwertfeger, Chelsea Finn, and Hang Zhao.
\newblock Robot parkour learning.
\newblock \emph{arXiv preprint arXiv:2309.05665}, 2023.

\bibitem[Zhuang et~al.(2024)Zhuang, Yao, and Zhao]{zhuang2024humanoid}
Ziwen Zhuang, Shenzhe Yao, and Hang Zhao.
\newblock Humanoid parkour learning.
\newblock \emph{arXiv preprint arXiv:2406.10759}, 2024.

\end{thebibliography}
}

\clearpage
\setcounter{page}{1}
\setcounter{section}{0}        
\setcounter{subsection}{0}     
\setcounter{subsubsection}{0}  
\maketitlesupplementary

\section{Real-world deployment Details}
\label{supp_sec:deploy}
\begin{figure}[!ht]
    \vspace{-5pt}
    \centering
    \includegraphics[width=0.9\linewidth]{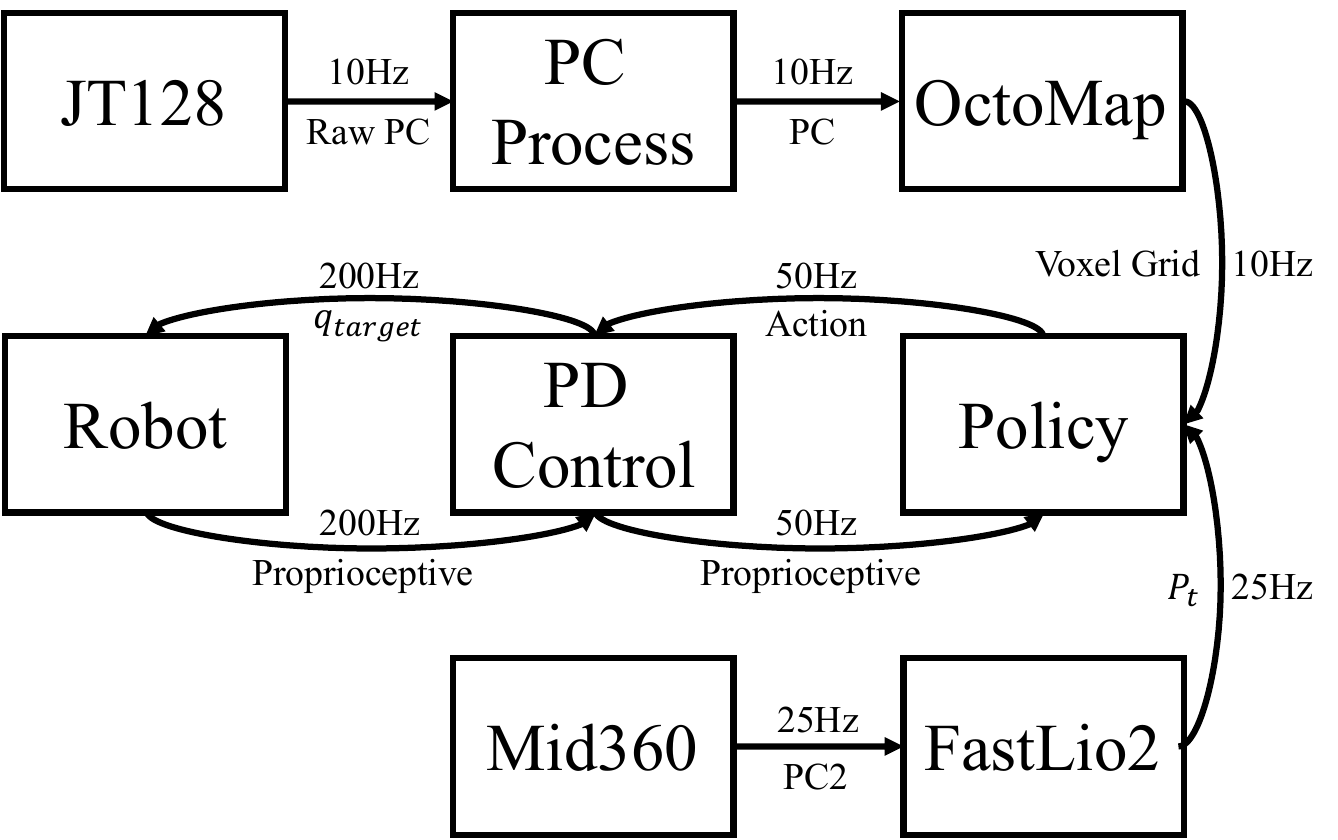}
    \caption{Diagram of information communication.}
    \label{fig:info}
    \vspace{-20pt}
\end{figure}
\paragraph{Target Position Command}
We use a Unitree G1 robot equipped with a Livox Mid360 LiDAR mounted on its head to run FastLIO2. The Mid360 is installed in a downward-facing orientation and provides a field of view of 360° horizontally and from $-7^\circ$ to $52^\circ$ vertically. The system provides the robot's position in the world coordinate frame at a frequency of 25\,Hz. To match this in simulation, the observation frequency of $P_t$ during training is also set to 25\,Hz. To align with our training setup, we initialize the robot's starting position at $(0, 0)$ and set the goal position for each run to $(4, 0)$. At each time step, FastLIO2 outputs the current position of the robot $(x, y)$, and the observation relative to the goal is defined as: $P_t = (4, 0) - (x, y) = (4 - x, -y)$.

\paragraph{Voxel Grid Processing}
We use two Hesai JT128 LiDARs mounted at the front and rear of the robot to collect raw point cloud data, which are merged and used for voxel grid construction. The JT128 supports 10\,Hz and 20\,Hz output modes; empirical testing showed that 20\,Hz leads to lower point cloud quality and reduced policy success rates. Therefore, we adopt the more reliable 10\,Hz mode and align the simulation accordingly. Each JT128 provides a vertical field of view of approximately 95°, a full 360° horizontal view, and 128 channels. The dual-sensor setup ensures near-complete coverage around the robot. To improve voxel grid quality, the raw point clouds are processed with the Octomap~\cite{hornung13auro} before being passed to the policy. In practice, using Octomap consistently leads to better performance.

\paragraph{Information Communication} Our system is fully deployed on a Unitree G1 robot using only an NVIDIA Orin NX, which has limited communication performance. When using TCP to transmit LiDAR data and LCM for internal robot state sharing, we observed a delay of approximately 200\,ms in proprioceptive data transmission, which is unacceptable. Thus, we made the following adjustments:
\begin{itemize}
    \item LiDAR output is clipped to include only points within the voxel grid used for perception, reducing data size.
    \item The voxel grid from Octomap and that used for observation share memory to avoid redundant transmission.
    \item Robot state reading and action command delivery are also implemented via shared memory, bypassing LCM.
\end{itemize}

These optimizations eliminate nearly all communication induced latency, except for inherent sensor delays. Overall information communication process is shown in ~\cref{fig:info}.

\section{Training Details}
\paragraph{Hyperparameter}
Our training framework is derived from ~\cite{xu2025facet}, and below is a summary of key PPO hyperparameters used in the training process:
\begin{table}[h]
\centering
\small
\begin{tabular}{l c}
\toprule
\textbf{Hyperparameter} & \textbf{Value} \\
\midrule
Environment number & $1024 \times 8$ \\
Steps per iteration 
PPO epochs & 4 \\
Minibatches & 8 \\
Clip range & 0.2 \\
Entropy coefficient  & 0.003 \\
GAE factor $\lambda$ & 0.95 \\
Discount factor $\gamma$ & 0.99 \\
Learning rate & $5e^{-4}$ \\
\bottomrule
\end{tabular}
\caption{Hyperparameters and their values.}
\label{tab:hyper}
\end{table}
\vspace{-10pt}
\paragraph{Policy Network Structure}
The Actor and Critic in our policy share the same network structure but maintain separate parameters. The shared architecture is illustrated in the block diagram below. To be specific, a two-layer MLP with hidden dimensions of 256 is used to encode non-voxel information (e.g., proprioceptive input). Note that the Critic additionally receives privileged observations, resulting in a slightly higher input dimension. This produces an intermediate feature $h_{\text{mlp}}$. In parallel, a three-layer 2D CNN processes the voxel grid input, producing a feature vector $h_{\text{cnn}}$. The two features are concatenated and passed through another MLP to produce a 256-dimensional latent representation. This latent vector is then fed into a final MLP:
\begin{itemize}
    \item The Actor outputs an action vector of dimension 29.
    \item The Critic outputs a scalar value estimate.
\end{itemize}
We use the Mish activation function throughout all layers.
\vspace{-5pt}
\[
\boxed{%
\begin{aligned}
\text{MLP:}\quad
    &h_{\text{mlp}}^{(1)} = \mathrm{Mish}\!\big(\mathrm{LN}(W_{\text{mlp},1} x_{\text{mlp}} + b_{\text{mlp},1})\big) \\
    &h_{\text{mlp}} = W_{\text{mlp},2} h_{\text{mlp}}^{(1)} + b_{\text{mlp},2},
      \quad \dim(h_{\text{mlp}})=256 \\[4pt]
\text{CNN:}\quad
    &z_1 = \mathrm{Mish}\!\big(\mathrm{Conv}(x_{\text{cnn}}; C{=}8, k{=}3, s{=}2, p{=}1)\big) \\
    &z_2 = \mathrm{Mish}\!\big(\mathrm{Conv}(z_1; C{=}8, k{=}3, s{=}2, p{=}1)\big) \\
    &z_3 = \mathrm{Mish}\!\big(\mathrm{Conv}(z_2; C{=}8, k{=}3, s{=}2, p{=}1)\big) \\
    &h_{\text{cnn}}^{\text{flat}} = \mathrm{Flatten}(z_3) \\
    &h_{\text{cnn}}^{(1)} = \mathrm{Mish}\!\big(\mathrm{LN}(W_{\text{cnn},1} h_{\text{cnn}}^{\text{flat}} + b_{\text{cnn},1})\big) \\
    &h_{\text{cnn}} = W_{\text{cnn},2} h_{\text{cnn}}^{(1)} + b_{\text{cnn},2},
      \quad \dim(h_{\text{cnn}})=64 \\[4pt]
\text{Fusion:}\quad
    &f = [h_{\text{mlp}},\, h_{\text{cnn}}] \\
    &h_{\text{out}}^{(1)} = \mathrm{Mish}(f) \\
    &h_{\text{out}} = \mathrm{Mish}\!\big(W_{\text{out}} h_{\text{out}}^{(1)} + b_{\text{out}}\big),
      \quad \dim(h_{\text{out}})=256
\end{aligned}
}
\]

\paragraph{Observation}
The composition of the observation is detailed in \cref{sec:method}, and the dimensionality of each observation component at a single time step $t$ is summarized in \cref{tab:obsdim}. The dimension of $Height\_Map_t$ shown in \cref{tab:obsdim} corresponds to its flattened form. Before flattening, it is represented as a $33 \times 33$ tensor. Specifically, this map captures the local terrain height around the robot, centered at its base, over a rectangular area with $x \in [-0.8, 0.8]$\,m and $y \in [-0.8, 0.8]$\,m. A resolution of $0.05m$ is used along both axes, resulting in one height ($z$) sample per $(x, y)$ grid point. This resolution is consistent with that used in the voxel grid.
Instead of applying fixed scaling, the observations are processed using a trainable \texttt{vecnorm} module before being fed into the policy. Vecnorm is applied in both training and deployment.
\begin{table}[h]
\centering
\small
\begin{tabular}{l c}
\toprule
\textbf{Observation Term} & \textbf{Dimension} \\
\midrule
$P_t$ & 4 \\
$T_{elapse,t}$ & 1 \\
$T_{left,t}$ & 1 \\
$a_t$ & 29 \\
$\omega_t$ & 3 \\
$g_t$ & 3 \\
$q_t$ & 29 \\
$\dot{q}_t$ & 29 \\
$Voxel\_Grid_t$ & $[32\times 32\times 40]$ \\
$v_t$ & 3 \\
$Height\_Map_t$ & 1089\\
\bottomrule
\end{tabular}
\caption{Observation terms and their dimensions.}
\label{tab:obsdim}
\end{table}

\paragraph{Reward}
Most reward components used in Gallant follow \citet{ben2025homie}, with necessary modifications to support our target-based formulation. In addition to the sparse target-reaching reward $r_{\text{reach}}$ introduced in \cref{sec:method}, we incorporate auxiliary shaping terms to improve sample efficiency during early training, as suggested by \citet{rudin2022advanced}.

We design the following three general-purpose rewards to encourage effective behavior across a variety of terrain conditions:

\begin{itemize}
    \item \textbf{Directional velocity reward:}
    \[
    r_{\text{velocity\_direction}} = \frac{\mathbf{a}(\mathbf{p}, \mathbf{g}) \cdot \mathbf{v}_t}{\|\mathbf{a}(\mathbf{p}, \mathbf{g}) \cdot \mathbf{v}_t\|_2},
    \]
    where $\mathbf{v}_t$ is the robot's instantaneous velocity and $\mathbf{a}(\mathbf{p}, \mathbf{g})$ is a direction vector incorporating both goal alignment and obstacle avoidance. It is computed as:
    \[
    \mathbf{a}(\mathbf{p},\mathbf{g}) = \sum_{j\in\mathcal{N}(\mathbf{p},r)} w_j\,\mathbf{u}_{r,j}
    + \kappa \sum_{j\in\mathcal{N}(\mathbf{p},r)} w_j\,\gamma_j\,\mathbf{t}_j,
    \]
    where $\mathcal{N}(\mathbf{p},r)$ denotes obstacle points within radius $r=1$\,m from the robot position $\mathbf{p}$; $\mathbf{u}_{r,j}$ is the repulsion unit vector from obstacle $j$ to the robot; $\mathbf{t}_j$ is a tangential unit vector (left/right) around obstacle $j$;
    \[
    w_j = \frac{\left[\max\left(1 - \frac{\max(d_j - 0.2,\, 0.02)}{0.8},\, 0\right)\right]^2}{\max(d_j - 0.2,\, 0.02)}
    \]
    is a distance-based weighting factor, and $\gamma_j = \max(\mathbf{g}^\top \mathbf{d}_j,\, 0)$ filters obstacles behind the goal direction. We set $\kappa = 0.8$ to weight the tangential term. This direction computation is only applied to relevant structures such as cylinders in \textit{Forest} and walls in \textit{Door}, and is efficiently parallelized via \texttt{warp}.

    \item \textbf{Head height reward:}
    \[
    r_{\text{head\_height}} = \exp\big(-4 (H_{\text{head\_est}} - H_{\text{head}})^2\big),
    \]
    where $H_{\text{head\_est}}$ is computed by shifting the robot 0.45\,m forward along the direction to the goal, averaging the terrain height within a $0.5 \times 0.5$\,m square, and subtracting a 0.1\,m offset. This reward encourages the robot to proactively lower its head to pass under overhead obstacles like ceilings.

    \item \textbf{Foot clearance reward:}
    \[
    r_{\text{feet\_clearance}} = \exp\big(-4 (H_{\text{feet\_est}} - H_{\text{feet}})^2\big),
    \]
    where $H_{\text{feet\_est}}$ is calculated similarly by querying terrain 0.5\,m ahead of each foot and averaging the height in a square region. Unlike \citet{ben2025homie}, who use terrain height directly under the foot, our design promotes proactive leg lifting over steps or platforms.
\end{itemize}

All three rewards are geometry-aware and general-purpose. They are computed consistently across all terrains without task-specific tuning and significantly improve the robot's ability to traverse diverse obstacle configurations.

\begin{figure*}[!ht]
    \vspace{-5pt}
    \centering
    \includegraphics[width=1.0\linewidth]{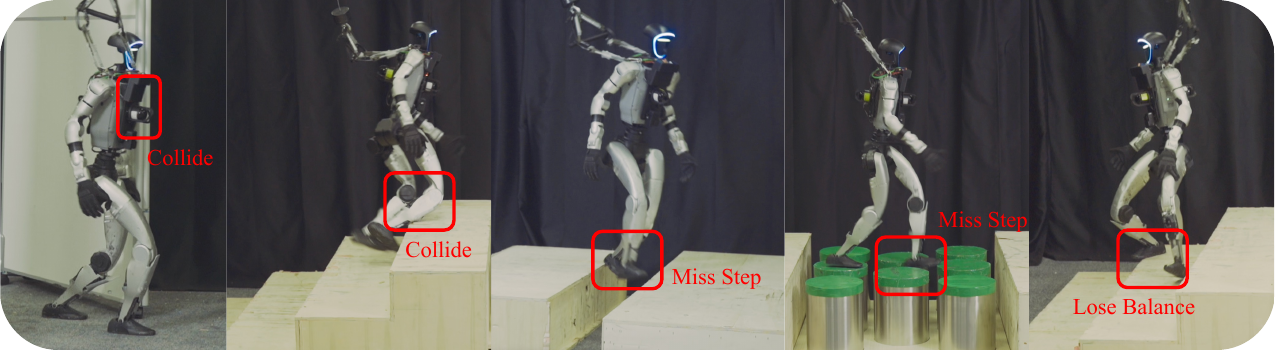}
    \caption{Failure Mode of policy trained by Gallant without LiDAR-related domain randomization.}
    \label{fig:fail}
    \vspace{-10pt}
\end{figure*}

\paragraph{Domain Randomization}
In addition to the LiDAR-specific domain randomization described in \cref{sec:method_lidar}, we apply several general randomization strategies during training to improve policy robustness:

\begin{itemize}
    \item \textbf{Mass randomization:} The masses of the pelvis and torso links are randomized as $m_{\text{rand}} = m \times \mathbf{U}(0.8, 1.2)$, where $\mathbf{U}$ denotes a uniform distribution.
    
    \item \textbf{Foot-ground contact randomization:} While the ground friction coefficient is fixed at 1.0, the foot joint friction is sampled from $\mathbf{U}(0.5, 2.0)$, and the restitution coefficient from $\mathbf{U}(0.05, 0.4)$.
    
    \item \textbf{Control parameter randomization:} The joint stiffness and damping parameters are randomized as $
    K_{p,\text{rand}} = K_p \times \mathbf{U}(0.8, 1.2), 
    K_{d,\text{rand}} = K_d \times \mathbf{U}(0.8, 1.2),
    $
    where $K_p$ and $K_d$ follow the settings in \citet{liao2025beyondmimic}.
    
    \item \textbf{Torso center-of-mass offset:} The center of mass position of the torso is perturbed by an offset sampled from $\mathbf{U}(-0.05, 0.05)$ along each axis.
    \item \textbf{Init Joint Position offset:} A random offset sampled from $\mathbf{U}(-0.1, 0.1)$ is also added to the robot's default joint positions and default joint velocities (0\,rad/s). This perturbation is applied during environment reset to randomize the robot's initial state.
\end{itemize}

\paragraph{Termination}
We apply several termination conditions during training to encourage effective and safe behavior:

\begin{itemize}
    \item \textbf{Force contact:} If any external force acting on the torso, hip, or knee joints exceeds 100\,N at any timestep, the episode is terminated.
    
    \item \textbf{Pillar fall:} For pillar-based terrains, if a foot penetrates more than 10\,cm below the ground level, the episode is terminated to prevent the robot from bypassing the obstacle by jumping off.
    
    \item \textbf{No movement:} To prevent the agent from exploiting reward shaping by staying on intermediate platforms, the episode is terminated if the robot fails to move at least 1\,m away from its initial position within 4 seconds.
    
    \item \textbf{Fall over:} The episode terminates when the robot loses balance and falls.
    
    \item \textbf{Feet too close:} Since self-collision is disabled during training to speed up simulation, this condition prevents the robot's feet from crossing or overlapping unnaturally.
\end{itemize}

\paragraph{Symmetry}
Following \citet{ben2025homie}, we apply symmetry-based data augmentation to accelerate training. In addition to flipping the proprioceptive observations as in their method, we also apply a flip along the $y$-axis to the perception representation. Specifically, the $(32, 32, 40)$ grid map is mirrored along the $y$ dimension to align with the flipped proprioceptive input, forming a consistent flipped observation. The reward remains unchanged under the transformation. Both original and flipped samples are stored together in the rollout buffer and jointly used during training.

\section{Failure Mode}
In \cref{fig:fail}, we illustrate typical failure cases of the NoDR variant (Gallant without domain randomization), as discussed in \cref{sec:real_aba}. These failures fall into three main categories as listed below:

\begin{itemize}
    \item \textbf{Latency-induced collision:} Due to sensor latency, the robot perceives a voxel grid that reflects the environment state from 100--200\,ms earlier. Since the policy is trained in simulation with instantaneous observations, it fails to react proactively and collides with obstacles it believes to be farther away.
    
    \item \textbf{Missed gap detection:} The robot occasionally fails to detect gaps in time, resulting in missed steps. This issue is particularly pronounced in scenarios like the \textit{Platform} task, where long-range gap perception is essential, leading to a lower success rate for NoDR.
    
    \item \textbf{Poor state estimation:} The robot exhibits imprecise estimation of its own body state. While it may avoid collisions or missed steps on stairs, it still enters unstable configurations and loses balance.
\end{itemize}

These observations highlight the importance of domain randomization, especially in simulating LiDAR latency and noise. Without such randomization, the policy fails to generalize effectively to real-world deployments.

\end{document}